\documentclass{article}

\usepackage{arxiv}

\usepackage[utf8]{inputenc} 
\usepackage[T1]{fontenc}    
\usepackage{hyperref}       
\usepackage{url}            
\usepackage{booktabs}       
\usepackage{amsfonts}       
\usepackage{nicefrac}       
\usepackage{microtype}      
\usepackage{lipsum}
\usepackage{graphicx}
\usepackage[ruled,linesnumbered]{algorithm2e}
\usepackage{amsmath}
\usepackage{url}
\usepackage{setspace}
\usepackage{multirow}
\graphicspath{ {./images/} }

\title{A Comprehensive Survey on the Ambulance Routing and Location Problems}

\author{
 Joseph Tassone \\
  Department of Computer Science\\
  Lakehead University\\
  Thunder Bay, ON, P7B 5E1 \\
  \texttt{jtasson2@lakeheadu.com} \\
  \And
 Salimur Choudhury \\
  Department of Computer Science\\
  Lakehead University\\
  Thunder Bay, ON, P7B 5E1 \\
}

\begin{document}
\maketitle
\begin{abstract}
In this research, an extensive literature review was performed on the recent developments of the ambulance routing problem (ARP) and ambulance location problem (ALP). Both are respective modifications of the vehicle routing problem (VRP) and maximum covering problem (MCP), with modifications to objective functions and constraints. Although alike, a key distinction is emergency service systems (EMS) are considered critical and the optimization of these has become all the more important as a result. Similar to their parent problems, these are NP-hard and must resort to approximations if the space size is too large. Much of the current work has simply been on modifying existing systems through simulation to achieve a more acceptable result. There has been attempts towards using meta-heuristics, though practical experimentation is lacking when compared to VRP or MCP. The contributions of this work are a comprehensive survey of current methodologies, summarized models, and suggested future improvements.
\end{abstract}

\section{Introduction}
"Emergency Medical Service" (EMS) systems are complex and fraught with problems requiring resolution. In these scenarios, two important questions are where to place the services and how to route them. Specifically these two problems are generalizations of the "Vehicle Routing Problem" (VRP) \cite{toth2002vehicle} and "Maximum Coverage Problem" (MCP) \cite{10.1007/3-540-48777-8_2}; dubbed themselves as the "Ambulance Routing Problem" (ARP) \cite{8370582} and "Ambulance Location Problem" (ALP) \cite{loc_sum}. Like the problems which they are based on, the two are NP-hard and subject to a number of constraints and potential decisions. VRP and MCP have both been explored extensively in past research and a number of approaches have already been surveyed \cite{vrp1,vrp2,mcp1}. Both can be viewed in terms of a graph problem $G=(V,E)$, where $V=\{0,1,...,n\}$ is the node set and $E$ the route set between facilities or destinations \cite{vrpbook,Essa2016}.

\subsection{Vehicle Routing Problem}
\begin{figure}[ht]
  \centering
  \includegraphics[width=0.75\linewidth]{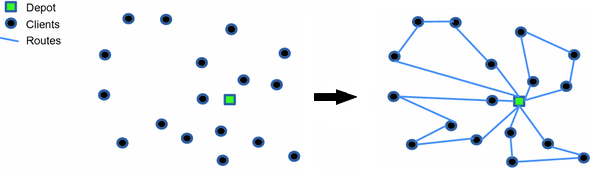}
  \caption{An expression of the basic vehicle routing problem with a single depot.}
  \label{vrp_diagram}
\end{figure}

In the VRP, the goal is to establish the optimal route (or set of routes) in which a fleet of vehicles can traverse to a set of clients or customers. Formally, there are a fixed set of vehicles with a centralized distribution depot. All vehicles must participate in deliveries to clients, with each one stating a certain demand or requirement. The goal is to organize the vehicles a way that minimizes the cost of routing \cite{toth2002vehicle}. There are many variations of the problem although almost all have common objectives of minimizing distance or cost in an optimal time window, while coordinating to customers and a depot (see Figure \ref{vrp_diagram}). Specific variations are based on limiting the vehicle capacity, implementing time windows for deliveries, allowing for multiple routes by a single vehicle, allowing for multiple trips, or providing additional services \cite{vrp_models_solutions,vrp_variants,vrp_backhaul,ELSHERBENY2010123,FAN20115284,LI20072918,multi_trip}. A diagram of the different common VRPs and their connections can be seen in Figure \ref{vrp_diagram}. Depending on the size and complexity the VRP can be modelled and solved with mathematical programming, though the NP-hard nature of the problem limits the use. It is much more effective to utilize metaheuristic approaches to gain a near optimal solution \cite{VERBIEST2018979}. Some of these solutions include "Genetic Algorithms" (GA), "Tabu Search", "Nearest Neighbor Search" (NNS), "Simulated Annealing", "Ant Colony Optimization" (ACO), and "Particle Swarm Optimization" (PSO) \cite{GA_VRP,HO20041947,Sarwono_2017,WEI2018843,BELL200441,5287655}. 

\begin{figure}[ht]
  \centering
  \includegraphics[width=0.65\linewidth]{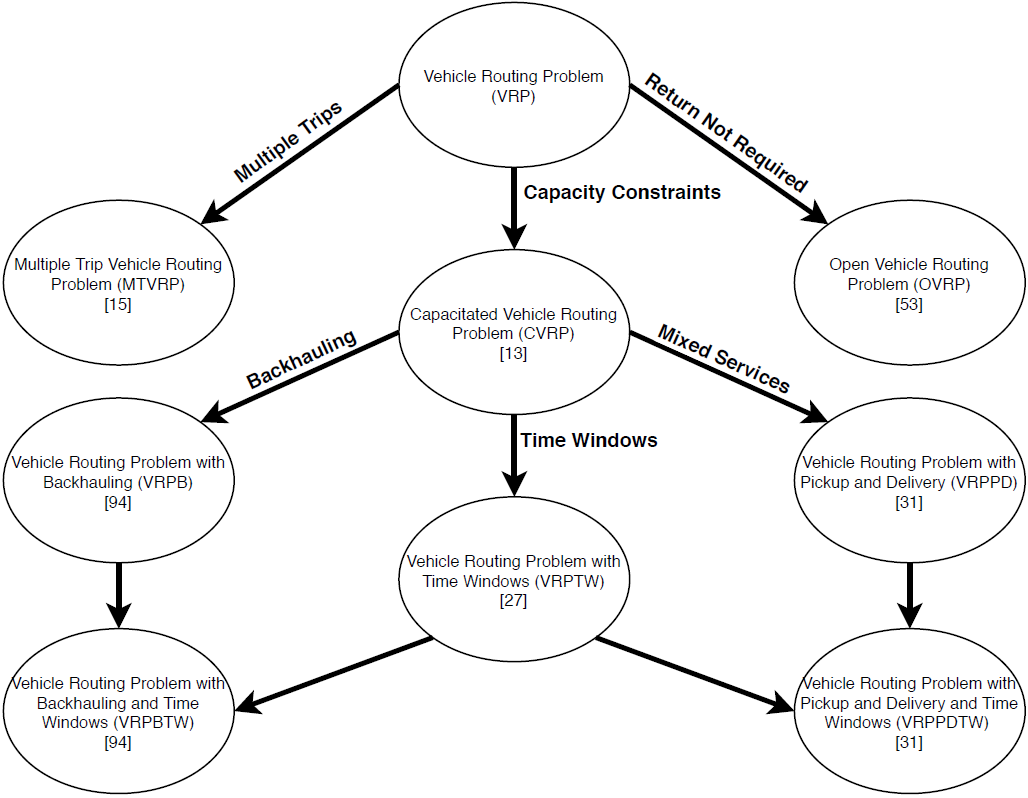}
  \caption{Common Vehicle Routing Problems and their Relationships.}
  \label{vrp_models_solutions}
\end{figure}

\subsection{Maximum Coverage Problem}
\begin{figure}[ht]
  \centering
  \includegraphics[width=0.6\linewidth]{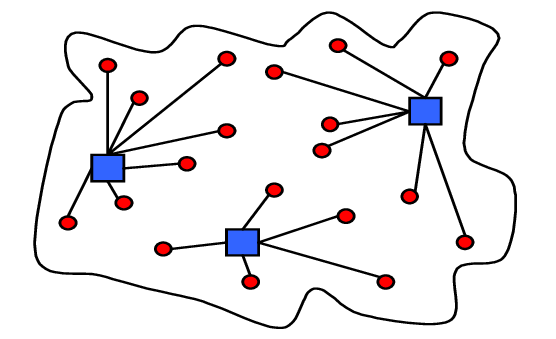}
  \caption{A representation of the the $k$-max facility location problem \cite{flp_citation}.}
  \label{flp}
\end{figure}

The MCP is a more specific instance of the "Set Coverage Problem", as one must choose at most $k$ sets which will cover a maximum number of elements \cite{10.1007/3-540-48777-8_2}. Similar to VRP, there are variations; the most documented extensions being being associated with weighted elements and placing limitations on cost. For the purposes of this study, a similar corresponding example is the facility location problem \cite{CHAUHAN20191}. There are at most $k$ facilities that must provide coverage to a certain number of customers (see Figure \ref{flp}). The customers within the range of the facility are assigned to it and are considered covered by the location. Each facility must maximize its reach and provide the greatest coverage, while adhering to the objective. Actual delivery to these customers becomes a secondary problem, which in itself may be solved as a VRP. Regardless, this can also be formulated in terms of integer linear programming (ILP), yet smaller cases can get an exact solution with algorithms like branch-and-bound \cite{Caprara2000}. In the case of larger examples, acceptable solutions can be gained by local-search, GA, or PSO \cite{10.1007/11970125_23,Atta2018,Balaji:2016:NAS:2993929.2993938}.

\subsection{Ambulance Routing and Location Problems}
As mentioned, both ARP and ALP are generalizations of the previously described problems. The ARP's goal is to determine the most effective routes for ambulances for emergency requests in disaster response situations or preplanned missions \cite{doi:10.1287/inte.2013.0683}. In this problem there is an ambulance set and a patient set. There is a cost associated with routing a patient with an ambulance and the goal is to minimize the total of all the assignments. For the ALP, it is primarily about determining the most effective location of ambulances for serving a population. It can be applied to either the facilities themselves or if there are a greater number of facilities than vehicles, such as in an air ambulance system, the placement of the vehicles among the facilities. Similar to the MCP, there are a set of facilities and a set of pickup points (patients) \cite{loc_sum}. Based on this setup, the process must maximize the coverage provided by the ambulance facilities to all patient pickup locations. This is only one example of a definition, as this problem has a multiple variations. Additionally, there is the potential overlap between the two cases where they are interrelated and solved simultaneously. In this situation the locations of ambulances are either based on the optimal routes or the optimal coverage is determined prior to routing \cite{BERLIN1974323}. Both problems can be modelled with integer-programming and either achieve an exact or approximate metaheuristic solution. The type varies greatly on the problem size, which is further based primarily on the area in which the ambulances are covering. A city wide EMS system will generally be varied and may be easily solved, while an air ambulance system may require increased consideration \cite{10.1007/978-3-030-04070-3_17}. Timing is also a far greater factor, as EMS is a critical system requiring fast response and decision making. For this reason exact solutions become difficult to justify, emphasizing a greater need for approximate metaheuristic variations. 

The vast majority of ARP and ALP are static, where either the locations never change or the planned missions are predetermined \cite{7731672}. Solutions to these problems are often generated with actual mission data and resolved for a specific period of time. However, a live EMS system generally considers real-time rescheduling, as disasters can occur at any point. In this case a dynamic solution must be developed for unexpected events or potential changes in the medical requirements of a patient \cite{SCHMID2012611}. To be clear, this does not suggest that all services must be dynamic and in some cases it may be completely unnecessary. A service may plan a day of missions without the expectation of change; although such a case is very specific and more likely to occur in a patient transfer scenario. This is not a completely new area of research as past dynamic solutions have been explored in the VRP \cite{GHIANI20031,PILLAC20131,Montemanni2005}. 

The remainder of this survey is arranged as follows. Section 2 discusses the modelling, constraints, and decision variables required for the ARP and ALP. Section 3 discusses potential solutions to the described problems, with descriptions of past related works. Section 4 discusses possible future research areas and suggests where further study should be explored. Finally, section 5 closes the paper providing a conclusion to the discussed topic.


\section{Process and Modelling}
As previously discussed, both the ARP and ALP have multiple forms conditional on the requirements and structure of the EMS system. Given that there are multiple versions of the problem, there are several ways to model the scenarios. All can be structured with integer-programming and solved subject to a set of constraints. While it is impossible to discuss all possible models, there are a few generalizations that can be made to each. This research only discusses the models which are most commonly used for resolving the ARP and ALP. This does not restrict the possibility of using other models, although the ones discussed are the basis for the majority seen currently. It is important to note that there are quite a few other variants and further research can be towards applying other VRP or MCP models.

\subsection{Ambulance Routing Problem}
Ambulance routing is highly connected to the location problem. The majority of ARP cannot actually be seen as full-scale VRP since the routes are usually a single pickup and drop-off for each scenario. While a general VRP usually has an entire route planned, an ambulance will usually return to the starting base following a drop-off. Exceptions exist in services such as air ambulance systems \cite{doi:10.1287/inte.2013.0683}; however, the majority do not follow this trend. As such, solving an ALP is sometimes more useful than solving an ARP. To be clear, solving an ALP will help to determine to optimal routing points and may cause the routing problem to be negligible. That being said, this is not an absolute and there are cases where ambulances may end up at new bases or perform intermediate stops based on scheduling. This is especially typical in air ambulance systems where vehicles are limited, patient transfers are common, and the coverage area is much larger \cite{doi:10.1287/inte.2013.0683}. The following is a simplified generalization of the ARP:

\begin{align}
    & \text{min}
    & &\sum_{i\in I}\sum_{j\in \mathcal{P}(J)}c_{ij}x_{ij} \\
    & \text{subject to} 
    & & \sum_{i\in I}\sum_{j\in \mathcal{P}(J)} x_{ij}= 1 , \forall \:r \\
    &&& \sum_{j\in \mathcal{P}(J)} x_{ij}\:\leq \:1 , \forall \:i \\
    &&& x_{ij} \:\in \:\{0,1\} , \forall \:j
\end{align}

The objective of the above model is to minimize the cost of routing all requests while utilizing the available ambulances. In this scenario $J$ is the set of requests and $I$ are the set of ambulances. For this case there exists the possibility that multiple requests will be serviced by a single vehicle; therefore, the model considers $\mathcal{P}(J)$ as the powerset of $J$. This indicates that an individual request is represented by $r$, while an single subset of $\mathcal{P}(J)$ be shown as $j$. Additionally, $c_{ij}$ is the cost of serving request $i$ with ambulance $j$, while $x_{ij}$ is a binary variable for determining whether request $i$ is being served by ambulance $j$. Equation 2 ensures that every request is satisfied and covered. The second constraint represented by Equation 3, states that each ambulance services at most a single set of requests. The final requirement guaranteed by Equation 4 simply enforces that $x_{ij}$ be limited to a binary decision. It should be noted that this scenario can be further constrained and implies that requests are predetermined. This is not unusual in the case of patient transfers, although is limited in a dynamic or diverging environment.

\subsubsection{Capacitated ARP}
The simple ARP can be modified like the VRP for a different scenarios through additional constraints. If the problem becomes a capacitated ambulance routing problem, then the constant capacity $Q$ must be considered for every vehicle \cite{cap_vrp_def}. In this scenario it is assumed that all vehicles are of the same type and posses the same capacity. The demand $d$ for vehicle $i$ cannot exceed $Q$, shown in Equation 5. The sum of each demand must be less than or equal to the capacity of the ambulance allotted to the request in question. Additionally, the demand is assumed to be known for each instance and the objective remains the same. 

\begin{equation}
    \sum_{i\in I}d_{i}x_{ij} \:\leq \: Q , \forall \:j
\end{equation}

\subsubsection{ARP with Time Windows}
Similarly, constraints may be added for time windows, where a solution cannot be considered if it breaks the upper (late) or lower bound (waiting) of a certain frame, represented as $[a_{i}, b_{i}]$ \cite{ELSHERBENY2010123}. The formulation is also modified a bit as $i$ and $j$ are starting and end point nodes for a customer, while $k$ is an ambulance. In this variation $t_{ij}$ is the time, including the service given to the patient at pickup. Another decision variable $s_{ik}$ indicates the time ambulance $k$ begins servicing patient $i$. For Equation 6, whenever an ambulance $k$ goes from $i$ to $j$, the calculation of $s_{ik} + t_{ij}$ must be less than or equal to $s_{jk}$. Essentially, this formulates the relationship from one pickup to the next for the time window. The last constraint given by Equation 7 establishes and enforces the time window.

\begin{equation}
    x_{ijk}(s_{ik} + t_{ij} - s_{jk}) \:\leq \: 0, \forall \:i, j, k
\end{equation}
\begin{equation}
    a_{i} \:\leq \: s_{ik} \:\leq \: b_{i}, \forall \:i, k
\end{equation}

\subsection{Ambulance Location Problem}
 An MCP is a broad term, while an ALP is more specific and in some ways closer to the "facility location problem" in its simplest form \cite{dcm_figure}. Other than this, there are many variations of this problem. A comparison of some of these of the ALP models can be found in \cite{7731672} including: "Maximal Covering Location" Problem (MCLP), "Double Standard Model" (DSM), "Average Response Time Model" (ARTM), "Maximum Availability Problem" (MAP), "Maximum Expected Covering Problem" (MECP), and "Expected Response Time Model" (ERTM). While it is not completely comprehensive, the several variations are discussed with their specific requirements. The most basic version of the problem is nearly identical to the MCP, with the key difference of maximizing the sum based on the weighted demand points covered by a vehicle.  

The base model is relatively simplistic and can be seen as the foundation for most ALPs. That being said there are extensions that can be made depending on the specifics attempting to be solved. For instance, the DSM extends the MCLP by attempting to cover each demand location by at least two ambulances within a small radius $r$. Another variation of the problem seeks to minimize the response times (example: ARTM), transforming the problem from a maximization to minimization. These versions utilize travel times $t_{ij}$ which must be less than or equal to a target response time. Each difference modifies the constraints or adds further fixed variables, such as how busy the ambulances are at a certain period. Regardless, all models solving the ALP seek to optimize the EMS system's coverage capability. Other than than MCLP current research has concentrated on the DSM, MECP, ARTM, and ERTM models. Each one has been formulated in Equations 10 through 35, displaying very similar characteristics to the simple MCLP. 

\subsubsection{Maximal Covering Location Problem}
In this optimization model (expressed in Equations 8 through 12), $I$ represents the set of disaster or demand locations, while $J$ is the set of active or possible bases \cite{doi:10.1287/inte.2013.0683}. The specifics of the sets depends on the circumstances, although can be summarized as starting and pickup point. The actual importance of a location is given a weight $d_{i}$, which is maximized by the objective function (Equation 8). The integer variable $x_{j}$ indicates the number of vehicles which are at base $j$ and $y_{i}$ is a binary variable stating whether a demand location $i$ is covered by an ambulance. Per Equation 9, every demand location needs to be covered by an occupied base. Additionally, each base can have at most $p$ ambulances; constrained by Equation 10. With regards to Equations 11 and 12, the decision variables in this problem are limited to a binary value of 0 or 1.

\begin{align}
    & \text{max}
    & &\sum_{i\in I}d_{i}y_{i} \\
    & \text{subject to} 
    & & \sum_{j\in J_{i}}x_{j}\geq y_{i} , \forall \:i \\
    &&& \sum_{j\in J}x_{j} \:\leq \:p , \forall \:j\\
    &&& x_{j} \:\in \:\{0,1\} , \forall \:j \\
    &&& y_{i} \:\in \:\{0,1\} , \forall \:i
\end{align}

\subsubsection{Double Standard Model}
The DSM is expressed in Equations 13 through 21, where the objective function seeks to maximize the demand coverage within a radius by two ambulances \cite{DIBENE2017107}. $y_{i}^1$ verifies whether the demand is being covered by at least one ambulance within a small radius, and $y_{i}^2$ checks if it is covered by two. Equation 15 uses $\alpha$ as a fraction of the demand locations that must be covered within a small radius. For this model, all demand location must be covered by an ambulance (Equation 14) and each must be covered twice (Equation 16). Per Equation 18, a demand location cannot be covered by twice unless it has been covered at least once.

\begin{align}
    & \text{max}
    & &\sum_{i\in I}d_{i}y_{i}^2 \\
    & \text{subject to} 
    & & \sum_{j\in J_{i}^2}x_{j}\geq 1 , \forall \:i \\
    &&& \sum_{i\in I}d_{i}y_{i}^1 \:\geq\: \alpha\sum_{i\in I}d_{i} , \forall \:i\\
    &&& \sum_{j\in J_{i}^2}x_{j} \:\geq \:y_{i}^1 + y_{i}^2 , \forall \:i\\
    &&& \sum_{j\in J}x_{j} \:= \:p , \forall \:j\\
    &&& y_{i}^1 \:\leq \: y_{i}^2 , \forall \:i\\
    &&& x_{j} \:\leq \:p_{j} , \forall \:j \\
    &&& x_{j} \:\in \:\mathbb{N} , \forall \:j \\
    &&& y_{i}^1, y_{i}^2  \:\in \:\{0,1\} , \forall \:i
\end{align}

\subsubsection{Maximum Expected Covering Problem}
Equations 22 through 26 show the formulation of the MECP \cite{doi:10.1080/10170660709509061}. For all of the demand locations, this model performs maximization on the weighted expected coverage. Through this it also considers the probability that an ambulance is available within a certain response time using a busy fraction represented by $q$. Whether $k$ vehicles can cover $i$ demand is expressed by the binary variable $y_{ik}$. Equation 23 maintains that an occupied base must be able to cover a demand location. Equation 24 prevents the number of ambulances from exceeding $p$, while Equations 25 and 26 limit the decision variables to natural numbers or binary values.

\begin{align}
    & \text{max}
    & &\sum_{i\in I}\sum_{k = 1}^{p}d_{i}(1-q)^{k-1}y_{ik} \\
    & \text{subject to} 
    & & \sum_{j\in J_{i}}x_{j} \:\geq\: \sum_{k = 1}^{p}y_{ik}, \forall \:i\\
    &&& \sum_{j\in J}x_{j} \:\leq \:p , \forall \:j\\
    &&& x_{j} \:\in \:\mathbb{N} , \forall \:j \\
    &&& y_{ik} \:\in \:\{0,1\} , \forall \:i, k\:\in \:\{1,...,p\}
\end{align}

\subsubsection{Average Response Time Model}
The ARTM differs from previous discussed variations in that its a minimization problem that seeks to lower the average response time relative to the nearest base \cite{Lanzarone:2018:ROA:3320516.3320818}. Travel time between base $j$ to demand $i$ is indicated by $t_{ij}$, and calculated within the objective function. A new binary variable called $z_{ij}$ indicates whether a base $j$ is nearest to a $i$ demand point and for this problem a base is limited to a capacity of one ambulance (Equation 21). Equation 28 prevents all $z_{ij}$ from not being set for every demand location $i$.

\begin{align}
    & \text{min}
    & &\sum_{j\in J}\sum_{i\in I}d_{i}t_{ij}z_{ij} \\
    & \text{subject to} 
    & & \sum_{j\in J_{i}}z_{ij} \:=\: 1, \forall \:i\\
    &&& \sum_{j\in J}x_{j} \:\leq \:p , \forall \:j\\
    &&& x_{j} \:\geq \: z_{ij} , \forall \:i,j\\
    &&& x_{j} \:\in \:\{0,1\} , \forall \:j \\
    &&& z_{ij} \:\in \:\{0,1\} , \forall \:i,j
\end{align}

\subsubsection{Expected Response Time Model}
The ERTM is somewhat of an extension of the ARTM, where the availability of an ambulance is considered and the objective is to minimize the expected response time for all demand locations \cite{7731672}. The model introduces a binary variable called $z_{ijk}$ which determines each ranked ambulance by location to a demand location. A rank is related to how near an ambulance is to a demand location, with closer vehicles having a greater rank. The variable is one when the ambulance is the $k^{th}$ nearest to the demand point. The probability of servicing a demand location is described by the extension of $q^{p-1}$, where $p$ is the rank of the ambulance. All $z_{ij}$ must be set for every demand location $i$ and every ambulance $k$, as per Equation 31. The model utilizes similar constraints to prior models with only a few modifications. 

\begin{align}
    & \text{min}
    & &\sum_{j\in J}\sum_{i\in I}\sum_{k = 1}^{p-1}d_{i}t_{ji}(1-q)q^{k-1}z_{ijk} \\
    &&& + \sum_{j\in J}\sum_{i\in I}d_{i}t_{ji}q^{p-1}z_{ijp}\nonumber \\
    & \text{subject to} 
    & & \sum_{j\in J}z_{ijk} \:=\: 1, \forall \:i, k \in \{1,...,p\}\\
    &&& x_{j} \:\geq \sum_{k=1}^{p}z_{ijk} , \forall \:i, j\\
    &&& \sum_{j \in J}x_{j} \:\leq \: p , \forall \:j\\
    &&& x_{j} \:\in \: \mathbb{N}, \forall \:j \\
    &&& z_{ijk} \:\in \:\{0,1\} , \forall \:i,j,k \in \{1,...,p\}
\end{align}

\subsubsection{Dynamic Models}
The models discussed thus far have all been static and will be lacking in terms of an online setting. In a static environment the inputs are known, while in a dynamic environment at least part of the input is revealed at a later point. This is common in an ambulance system where incidents cannot be predetermined and occur in real-time. The static environment is effective in the case of patient transfer, yet fails if one considers the entire picture. For this subset the problem is more related to relocation, diversion, or changes in routing. A full discussion of this modification was explored by Schmid \cite{SCHMID2012611,SCHMID20101293}. In these works cost was still minimized, although decisions made at a point in time would consider the influence they had for a later point in the system. Essentially, the system constrained and evaluated based on the future impact. This allowed the model to potentially make faster responses and be prepared for changing scenarios. Work in this area is still lacking and has the potential for further study.

\section{Solutions and Methodologies}
Both the ALP and ARP have a number of solutions and techniques which may be used for modelling and resolution. As already discussed, the problems are generally first modelled based on their respective instance and data. Afterwards, an approach then either uses some form of heuristic or a simulation to achieve a solution. A summary of the current research for the ALP and ARP can be found in Table \ref{tab:works_arp} and Table \ref{tab:works_alp}. The remainder of this section offers solutions for various possible techniques and then provides current research implementing those methodologies.

\begin{table}[!ht]
\caption{Summary of ARP Techniques and Models}
\label{tab:works_arp}
\resizebox{\columnwidth}{!}{
\begin{tabular}{llll}
\hline
\textbf{Reference \& Section}                                            & \textbf{Model}                                                                            & \textbf{Technique}                                                                                                                                   & \textbf{Objective}                                                                                                                  \\ \hline
\cite{5429196} (Section \ref{sim_and_lin})                               & ARP                                                                                       & \begin{tabular}[c]{@{}l@{}}Simulation, \\ DES,\\ Markov Decision Process,\\ Approximate dynamic programming\end{tabular}                             & Minimize average response time                                                                                                      \\ \hline
\cite{Ni:2012:EBA:2429759.2429818} (Section \ref{sim_and_lin})           & ARP                                                                                       & \begin{tabular}[c]{@{}l@{}}Simulation,\\ Approximate dynamic programming,\\ Mathematical Programming (Gurobi)\end{tabular}                           & Minimize average response time                                                                                                      \\ \hline
\cite{8370579} (Section \ref{sim_and_lin})                               & Bi-objective ARP                                                                          & $\epsilon$-constraint method                                                                                                                         & \begin{tabular}[c]{@{}l@{}}Minimize service completion time,\\ Minimize setup cost\end{tabular}                                     \\ \hline
\cite{8370582} (Section \ref{aco_op})                                    & Dynamic ARP                                                                               & ACO                                                                                                                                                  & Minimize total time                                                                                                                 \\ \hline
\cite{ant_rout_am} (Section \ref{aco_op})                                & Capacitated ARP                                                                           & ACO                                                                                                                                                  & Minimize total time                                                                                                                 \\ \hline
\cite{TLILI20181585} (Section \ref{ga_op})                               & ARP                                                                                       & GA                                                                                                                                                   & Minimize total travel cost                                                                                                          \\ \hline
\cite{Fogue:2016:NPT:2972705.2972878} (Section \ref{ga_op})              & Capacitated ARP                                                                           & Novel NURA algorithm relying on GA                                                                                                                   & Minimize average waiting time                                                                                                       \\ \hline
\cite{Nafarrate:2011:DCA:2431518.2431667} (Section \ref{ga_op})          & \begin{tabular}[c]{@{}l@{}}ARP,\\Ambulance diversion\end{tabular}                         & \begin{tabular}[c]{@{}l@{}}GA,\\ DES\end{tabular}                                                                                                    & Minimize average waiting time                                                                                                       \\ \hline
\cite{TALARICO2015120} (Section \ref{local_op})                          & ARP                                                                                       & \begin{tabular}[c]{@{}l@{}}LNS,\\ Mathematical Programming (CPLEX)\end{tabular}                                                                      & Minimize total lateness                                                                                                             \\ \hline
\cite{karakaya_2016} (Section \ref{local_op})                            & Capacitated ARP                                                                           & \begin{tabular}[c]{@{}l@{}}GA,\\ NNS,\\ CVRP Benchmarks\end{tabular}                                                                                 & Minimize total travel distance                                                                                                      \\ \hline
\cite{10.1186/s12913-016-1727-5} (Section \ref{local_op})                & \begin{tabular}[c]{@{}l@{}}Capacitated ARP with Time Windows,\\ MD-H-DARP\end{tabular}    & \begin{tabular}[c]{@{}l@{}}Tabu Search,\\ Mathematical Programming (Fico Xpress)\end{tabular}                                                        & Minimize total operation time                                                                                                       \\ \hline
\cite{REPOUSSIS2016531} (Section \ref{local_op})                         & Bi-objective ARP                                                                          & \begin{tabular}[c]{@{}l@{}}Iterated Tabu Search,\\ Mathematical Programming (CPLEX)\end{tabular}                                                     & \begin{tabular}[c]{@{}l@{}}Minimize total response time,\\ Minimize total flow\end{tabular}                                         \\ \hline
\cite{Tlili:2017:SAS:3141923.3142198} (Section \ref{pso_op})             & \begin{tabular}[c]{@{}l@{}}Open ARP,\\ ARP with Pickup and Delivery\end{tabular}          & \begin{tabular}[c]{@{}l@{}}Cluster-first route-second (petal algorithm),\\ PSO,\\ GA\end{tabular}                                                    & Minimize total travel cost                                                                                                          \\ \hline
\cite{7919560} (Section \ref{cluster_op})                                & ARP                                                                                       & \begin{tabular}[c]{@{}l@{}}K-Mean clustering,\\ Weighted K-Mean clustering,\\ Density-Based clustering,\\ Directly Reachable clustering\end{tabular} & Minimize total travel distance                                                                                                      \\ \hline
\end{tabular}
}
\end{table}
\begin{table}[!ht]
\caption{Summary of ALP Techniques and Models}
\label{tab:works_alp}
\resizebox{\columnwidth}{!}{
\begin{tabular}{llll}
\hline
\textbf{Reference \& Section}                                            & \textbf{Model}                                                                            & \textbf{Technique}                                                                                                                                   & \textbf{Objective}                                                                                                                  \\ \hline
\cite{SCHMID2012611} (Section \ref{sim_and_lin})                         & Dynamic ALP                                                                               & \begin{tabular}[c]{@{}l@{}}Simulation,\\ Approximate dynamic programming\end{tabular}                                                                & Minimize average response time                                                                                                      \\ \hline
\cite{6163878} (Section \ref{sim_and_lin})                               & \begin{tabular}[c]{@{}l@{}}ALP,\\ TAZ\_OPT\end{tabular}                                   & \begin{tabular}[c]{@{}l@{}}Simulation, \\ Goal programming\end{tabular}                                                                              & \begin{tabular}[c]{@{}l@{}}Minimize total deviations,\\ Maximize the expected coverage,\\ Minimize busyness likelihood\end{tabular} \\ \hline
\cite{7851626} (Section \ref{sim_and_lin})                               & \begin{tabular}[c]{@{}l@{}}ALP,\\ DSM\end{tabular}                                        & \begin{tabular}[c]{@{}l@{}}Simulation, \\ Goal programming,\\ Mathematical Programming (CPLEX)\end{tabular}                                          & Maximize total demand covered                                                                                                       \\ \hline
\cite{6675685} (Section \ref{sim_and_lin})                               & Dynamic ALP                                                                               & \begin{tabular}[c]{@{}l@{}}Simulation, \\ Fuzzy simultaneous data envelopment analysis,\\ Mathematical Programming (CPLEX)\end{tabular}              & \begin{tabular}[c]{@{}l@{}}Minimize facility location cost,\\ Minimize sum inefficiency values\end{tabular}                         \\ \hline
\cite{Morohosi:2012:HSA:2429759.2429903} (Section \ref{sim_and_lin})     & \begin{tabular}[c]{@{}l@{}}ALP,\\ MECP\end{tabular}                                       & \begin{tabular}[c]{@{}l@{}}Simulation,\\ Hypercube\end{tabular}                                                                                      & Maximize total expected demand covered                                                                                              \\ \hline
\cite{Lee:2012:SIM:2429759.2429871} (Section \ref{sim_and_lin})          & \begin{tabular}[c]{@{}l@{}}ALP,\\ MECP\end{tabular}                                       & \begin{tabular}[c]{@{}l@{}}Simulation,\\ DES\end{tabular}                                                                                            & Maximize expected number of transported patients                                                                                    \\ \hline
\cite{DIBENE2017107} (Section \ref{sim_and_lin})                         & \begin{tabular}[c]{@{}l@{}}ALP,\\ DSM\end{tabular}                                        & \begin{tabular}[c]{@{}l@{}}Simulation,\\ Mathematical Programming (MOSEK)\end{tabular}                                                               & Maximize total demand covered                                                                                                       \\ \hline
\cite{Rislien2018ComparingPA} (Section \ref{sim_and_lin})                & \begin{tabular}[c]{@{}l@{}}ALP,\\ MCLP\end{tabular}                                       & \begin{tabular}[c]{@{}l@{}}Simulation,\\ Mathematical Programming (CPLEX)\end{tabular}                                                               & Maximize total demand covered                                                                                                       \\ \hline
\cite{10.1371/journal.pone.0215385} (Section \ref{sim_and_lin})          & \begin{tabular}[c]{@{}l@{}}ALP,\\ MECP\end{tabular}                                       & \begin{tabular}[c]{@{}l@{}}Simulation,\\ Mathematical Programming (CPLEX)\end{tabular}                                                               & Maximize total expected demand covered                                                                                              \\ \hline
\cite{Lanzarone:2018:ROA:3320516.3320818} (Section \ref{sim_and_lin})    & \begin{tabular}[c]{@{}l@{}}ALP,\\ ERTM\end{tabular}                                       & \begin{tabular}[c]{@{}l@{}}Simulation,\\ DES,\\ Recursive optimization-simulation approach,\\ Mathematical Programming (CPLEX)\end{tabular}          & Minimize expected response times                                                                                                    \\ \hline
\cite{optimal_loc} (Section \ref{sim_and_lin})                           & \begin{tabular}[c]{@{}l@{}}ALP,\\ MECP\end{tabular}                                       & \begin{tabular}[c]{@{}l@{}}Simulation,\\ DES,\\ Hypercube,\\ Mathematical Programming (Premium Solver Platform)\end{tabular}                                                 & Maximize total expected demand covered                                                                                              \\ \hline
\cite{6761400} (Section \ref{sim_and_lin})                               & \begin{tabular}[c]{@{}l@{}}ALP,\\ Stochastic programming model\end{tabular}               & \begin{tabular}[c]{@{}l@{}}Simulation,\\ Sample average approximation,\\ Mathematical Programming (CPLEX)\end{tabular}                               & Minimize total expected costs                                                                                                       \\ \hline
\cite{8574936} (Section \ref{sim_and_lin})                               & Two-stage robust ALP                                                                      & \begin{tabular}[c]{@{}l@{}}C\&CG method with approximation,\\ Mathematical Programming (CPLEX)\end{tabular}                                          & Maximize total relative coverage                                                                                                    \\ \hline
\cite{7831330} (Section \ref{aco_op})                                    & Dynamic ALP                                                                               & \begin{tabular}[c]{@{}l@{}}ACO hybridized by a guided local search,\\ DES\end{tabular}                                                               & Minimize total lateness                                                                                                             \\ \hline
\cite{IANNONI2009528} (Section \ref{ga_op})                              & Bi-objective ALP                                                                          & \begin{tabular}[c]{@{}l@{}}GA,\\ Hypercube,\\ DES\end{tabular}                                                                                       & \begin{tabular}[c]{@{}l@{}}Minimize average response time,\\ Minimize imbalance of workloads\end{tabular}                           \\ \hline
\cite{doi:10.1080/10170660709509061} (Section \ref{ga_op})               & \begin{tabular}[c]{@{}l@{}}Bi-objective ALP,\\ MECP,\\ DSM\end{tabular}                   & GA                                                                                                                                                   & \begin{tabular}[c]{@{}l@{}}Maximize demand covered twice,\\ Maximize the expected number of demands\end{tabular}                    \\ \hline
\cite{Sasaki_10.1186/1476-072X-9-4} (Section \ref{ga_op})                & \begin{tabular}[c]{@{}l@{}}ALP,\\ ARTM\end{tabular}                                       & Modified grouping GA                                                                                                                                 & Minimize average response time                                                                                                      \\ \hline
\cite{7257213} (Section \ref{ga_op})                                     & \begin{tabular}[c]{@{}l@{}}Continuous ALP,\\ Weber Facility Location Problem\end{tabular} & GA                                                                                                                                                   & Minimize average response time                                                                                                      \\ \hline
\cite{Benabdouallah:2017:CGA:3128128.3128136} (Section \ref{ga_op})      & ALP                                                                                       & \begin{tabular}[c]{@{}l@{}}ACO,\\ GA,\\ DES,\\ Mathematical Programming (CPLEX)\end{tabular}                                                         & Minimize total lateness                                                                                                             \\ \hline
\cite{SCHMID20101293} (Section \ref{local_op})                           & \begin{tabular}[c]{@{}l@{}}ALP,\\ Multi-period DSM\end{tabular}                           & \begin{tabular}[c]{@{}l@{}}VNS,\\ Mathematical Programming (CPLEX)\end{tabular}                                                                      & Maximize weighted sum for coverage                                                                                                  \\ \hline
\cite{GENDREAU199775} (Section \ref{local_op})                           & \begin{tabular}[c]{@{}l@{}}ALP,\\DSM\end{tabular}                                         & \begin{tabular}[c]{@{}l@{}}Tabu Search,\\ Mathematical Programming (CPLEX)\end{tabular}                                                              & Maximize weighted sum for coverage                                                                                                  \\ \hline
\cite{sched_max_cov} (Section \ref{local_op})                            & \begin{tabular}[c]{@{}l@{}}Bi-objective ALP,\\ MECP\end{tabular}                          & \begin{tabular}[c]{@{}l@{}}Tabu Search,\\ Mathematical Programming (CPLEX)\end{tabular}                                                              & \begin{tabular}[c]{@{}l@{}}Maximize minimum expected coverage,\\ Maximize aggregate expected coverage\end{tabular}                  \\ \hline
\cite{edcalpfa} (Section \ref{local_op})                                 & \begin{tabular}[c]{@{}l@{}}ALP,\\DSM\end{tabular}                                         & \begin{tabular}[c]{@{}l@{}}ACO,\\ Tabu Search\end{tabular}                                                                                           & Maximize total demand covered                                                                                                       \\ \hline
\cite{psw_mc} (Section \ref{pso_op})                                     & \begin{tabular}[c]{@{}l@{}}ALP,\\ MCLP\end{tabular}                                       & PSO                                                                                                                                                  & Maximize total demand covered                                                                                                       \\ \hline
\cite{Li:2015:LSA:2820783.2820876} (Section \ref{cluster_op})            & ALP                                                                                       & PAM-based refinement                                                                                                                                 & Minimize average travel time                                                                                                        \\ \hline
\end{tabular}
}
\end{table}

\subsection{Simulation and Mathematical Programming}
\label{sim_and_lin}
\subsubsection{Simulation Description}
Simulation is generally used as a method of capturing the complexity of system or showing its operations by formulating it as a mathematical representation. It can be utilized as optimization technique where the system is modified and compared against a real existing case \cite{BIERLAIRE20154}. Simulation involves several methodologies, many of which are used to model complex processes. For instance, "discrete-event simulation" (DES) is used to model a system as a sequence of events, where each occurs at a certain instant in time and represents a change in the overall system’s state \cite{SEAY201685}. "Monte Carlo simulations" are related to DES; however, utilize random number generators to add a layer of nondeterminism for modelling stochastic problems. They generate a random sample set of points based on each given input value. "Hypercube" is a form of sampling that takes this further and uses multidimensional distribution for the generation of randomized sample parameters \cite{Morohosi:2012:HSA:2429759.2429903}. It is connect to Monte Carlo, although the goal is to more evenly distribute the points across the possible values. "Markov decision processes" are another example and model decisions in a stochastic and sequential manner. In this technique there is an assumed finite number of states and actions which can be taken. The state is randomly changed in response to choices made in the environment, where the goal is to maximize long-term total reward \cite{LITTMAN20019240}. 


\subsubsection{Mathematical Programming Description}
A way of representing or resolving a simulated model is by representing it with "mathematical programming". If the variables to be described are restricted to integers then it can be formulated as an "integer linear programming" (ILP) problem. In this case, all the variables are locked to either binary or integer values, and the constraints and objective functions are linear. A case where where some of the decision variables are not restricted also exists, and is referred to as a "mixed-integer linear program" (MILP) \cite{POE2017173}. There are a number of ways to resolves these mathematical formulations, with optimization solvers being a common choice for achieving exact solutions. Some choices for optimizers include "CPLEX", "Gurobi", "Fico Xpress", and "MOSEK".  

\subsubsection{Goal Programming Description}
A more generalized version of ILP is known as "goal programming", where there can be the possibility of solving multiple and possibly conflicting objectives. This can be used to simulate a decision making process where multiple goals are resolved and solved in order to reach a solution \cite{TAMIZ1998569}. Similar to ILP, goal programming can achieve an exact solution, although the time required to achieve this may make it unreasonable to do so. A larger problem space drastically increases the time required to do so, making it unreasonable in a critical EMS system. 

\subsubsection{Dynamic Programming Description}
"Dynamic programming" is another method of not just representing, but also solving a model. It takes a divide and conquer approach of optimization, where the solution depends on solving previous subproblems. The difference in this case is where divide and conquer subproblems are solved independently, in dynamic programming smaller subproblems are solved to gain larger subproblems \cite{REVEILLAC201555}. Essentially, a complex problem is recursively broken down in simpler variations so that the whole can be solved. This is especially relevant in a decision making process, where an individual decision may be broken down into is respective steps.

\subsubsection{Simulation and Mathematical Programming Solutions}
\begin{figure}[ht]
  \centering
  \includegraphics[width=0.65\linewidth]{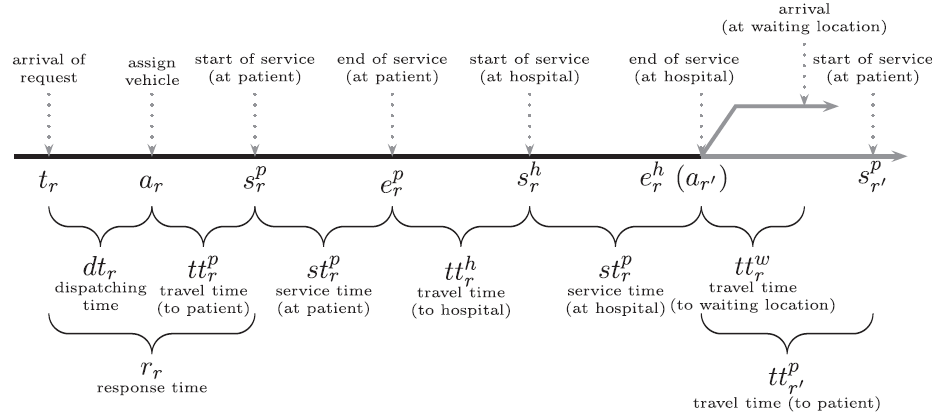}
  \caption{Full representation of dispatching and relocation.}
  \label{1_schmid}
\end{figure}

Schmid proposed a stochastic dynamic model for ambulance relocation and dispatching in Vienna, resolving it with approximate dynamic programming (ADP) \cite{SCHMID2012611}. A full description of the problem can be seen in Figure \ref{1_schmid}; whereby a request is received, a vehicle is assigned, a vehicle travels to the site, a vehicle travels to a hospital, and then returns to waiting. Other approaches based on dynamic programming cannot handle high-dimensional state spaces and generally have a difficult time dealing with issues revolving around the dimensionality in large scale stochastic problems. In this model, rather than stepping backward in time, the current strategy iteratively stepped forward and uses an approximate value function. In order to ensure dynamic decision making, they estimated the value of being in the current state and the states from the previous and corresponding decisions. The results of the experiments with real-world data outperformed current practices and decreased the average response time by 12.89\% (see Table \ref{adp_tab_results}).

\begin{table}[!ht]
\caption{Results of different policies performed with real data.}
\label{adp_tab_results}
\resizebox{0.75\columnwidth}{!}{\begin{tabular}{|l|l|l|l|}
\hline
\textbf{Policy} & \textbf{Average Response (min)} & \textbf{Minimum Response (min)} & \textbf{Maximum Response (min)} \\ \hline
Current         & 4.35                            & 3.73                            & 5.21                            \\ \hline
Random          & 5.04                            & 4.44                            & 5.65                            \\ \hline
ADP             & 4.04                            & 3.07                            & 5.02                            \\ \hline
\end{tabular}}
\end{table}

Maxwell et al. observed an ambulance redeployment scenario to reduce response times through the repositioning of idle vehicles \cite{5429196}. The operations of the EMS were simulated using a DES and evaluated in the context of approximate dynamic programming. The actual redeployment problem was designed as a Markov decision process and revealed that the simulation performed better than current static policies. In their formulation they determined that the time of a state could be very close to the subsequent one. As such, utilizing a one-step simulation would have given almost no information on choosing a particular action. For this reason they utilized micro-simulations, which were independent for each state and stopped only once a decision was reached. The micro-simulations did have an effect on the final evaluation of the system, which is shown and separated by the particular policy in Table \ref{maxwell_results}.

\begin{table}[!ht]
\caption{Runtime of each ADP policy based on the amount of micro simulations.}
\label{maxwell_results}
\resizebox{0.5\columnwidth}{!}{\begin{tabular}{l|l|l|l|l|l|}
\cline{2-6}
                                          & \multicolumn{5}{l|}{\textbf{Micro Simulations}}                    \\ \hline
\multicolumn{1}{|l|}{\textbf{ADP Policy}} & \textbf{5} & \textbf{10} & \textbf{30} & \textbf{60} & \textbf{10} \\ \hline
\multicolumn{1}{|l|}{Decision Time (s)}   & 0.013      & 0.025       & 0.077       & 0.152       & 0.253       \\ \hline
\multicolumn{1}{|l|}{Simulation Time (s)} & 15         & 30          & 89          & 177         & 289         \\ \hline
\multicolumn{1}{|l|}{Iteration Time (m)}  & 7.7        & 15          & 45          & 87          & 145         \\ \hline
\multicolumn{1}{|l|}{Training Time (h)}   & 3.2        & 6.3         & 19          & 39          & 60          \\ \hline
\end{tabular}}
\end{table}

Ni et al. viewed a real-time problem involving ambulance deployment, where simulation was used to devise a redeployment strategy in order to minimize response times \cite{Ni:2012:EBA:2429759.2429818}. A simulation was developed as a stochastic dynamic program (see Equation 39) where patient calls were serviced based on a "first-come-first-serve" method. In this $X(s)$ denotes the set of feasible actions, $U$ are random uniform variables held in a vector for capturing random noise, $c(s,x,U)$ denotes the respective decision cost, and the next state relative to the current state is given by $f(s,x,U)$ . In this model after pickup, an ambulance would take the patient to a hospital and after hospital transportation could potentially service another call. Essentially the bounds were obtained and explored from a combination of methodologies based on comparison in queues. The results of the simulation yielded possible improvements in the deployment strategies of the ambulances.

\begin{equation}
  V(s) = \min_{x\in X(s)} \mathbb{E}\left \{ c(s,x,U) + V(f(s,x,U)) \right \}
\end{equation}

Shuib and Zaharudin developed a custom model for maximizing the availability of ambulances \cite{6163878}. The model in question is known as "time-based ambulance zoning optimization" (TAZ\_OPT), a goal programming approach which reduced response times. The model's primary purpose was to use a grid-based system to identify the satellite location for ambulances, while also determining the specific number allocated to a location. In the model, they used the probability of an ambulance reaching a disaster location within a target range to determine its coverage and ensure it would only serve the specific area. Additionally, the model also optimized the future expected coverage based on the demand and determined the busyness fraction of an active ambulance. The formulation of the goal programming approach can be seen in Equation 40 whereby $P_{0}$ and $P_{1}$ are the first (Equation 41) and second (Equation 42) goals, and $d_{0}^{-}$ and $d_{1}^{+}$ are the over or under attainment of those respective goals. Respectively $di$ is proportion of daily demand rate in grid $i$, $P_{ij}$ is the probability of reaching grid $i$ in the target response time from location $j$, $Y_{ij}$ is a binary decision for whether an ambulance at location $j$ is the nearest location to $i$, $m$ is the total amount of grids, and $c$ is the maximum number of vehicles that be put towards grid $j$.

\bigskip

\begin{equation}
    \min (P_{0}d_{0}^{-}+P_{1}d_{1}^{+})
\end{equation}

\begin{equation}
    P_{0} =\max \sum_{i}^{n}d_{i}P_{ij}Y_{ij} \; \forall j = 1,2,...,m
\end{equation}

\begin{equation}
    P_{1} =\min(B_{j,k}) \; \forall j = 1,2,...,m ;\forall k = 1,2,...,c
\end{equation}

\smallskip

\begin{figure}[ht]
  \centering
  \includegraphics[width=0.6\linewidth]{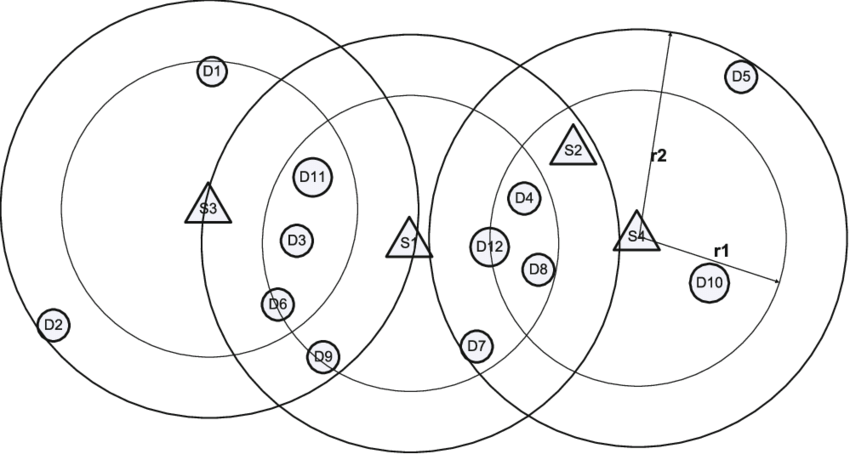}
  \caption{Graphical representation of the double coverage problem \cite{dcm_figure}.}
  \label{2_lahijanian}
\end{figure}

Lahijanian et al. considered a double coverage concept, where the demand was covered by a minimum of two vehicles \cite{7851626}. Figure \ref{2_lahijanian} shows a representation of this concept; where demand points are represented by circles and current vehicle locations are represented by triangles. In this scenario, demand points must be covered twice within $r1$, while secondary necessity is held within $r2$. Unlike prior works, they also considered the uncertainty of travel times between patient and ambulance locations with "triangular fuzzy numbers". To resolve the model, they took a goal programming approach and received a solution using "GAMS".

Khodaparasti and Maleki presented a new combined dynamic EMS model that sought to locate vehicles and stations for disaster situations \cite{6675685}. In this model, they considered uncertainty within the parameters and implemented a dynamic structure utilizing a fuzzy numbers for inputs and outputs. They obtained both dynamic locations and were also able to analyze the efficiency of the locations at the same time with data envelopment analysis. The advantage of this structure was allowing the ability to make short-term decisions, rather than considering long-term periods only. The objective function is unique and expressed by Equation 43, where it minimizes the cost of locating facilities. For this problem $I$ is the set of demand points, $J$ is the set of candidate ambulance stations, $T$ is the set of time periods, $S_{t}$ is the set of possible scenarios during period $t$, $f_{jt}$ is the operation costs of station $j$ for period $t$, $h_{ist}$ is amount of demand, $d_{ijst}$ is the distance from a demand point to a potential site, $\lambda$ is the minimum satisfaction degree of the membership function, and the binary decision variables are $x_{jt}$ (station) and $y_{jkst}$ (ambulance). In this function the first set of terms is related to station operation costs, the second is the cost of an assignment, and the last is the efficiencies of the stations.

\begin{equation}
    \min \sum_{t\in T}\sum_{j\in J}f_{jt}x_{jt} + \sum_{i\in I}\sum_{j\in J}\sum_{t\in T}\sum_{s\in S_{t}}\sum_{k=1}^{p}h_{ist}d_{ijst}y_{jkst}+(1-\lambda)
\end{equation}

Morohosi and Furuta analyzed the patterns in dispatching real ambulances and compared this with a simulated output \cite{Morohosi:2012:HSA:2429759.2429903}. Then based on this they attempted to utilize a simple methodology for estimating the improvement from an optimal solution location for Tokyo. For this purpose, they developed a hypercube simulation model which generated randomized parameter values and incorporated time nonhomogeneity. A key feature identified in their simulation was the introduction of a ambulance priority list for a given demand position, speeding the process up. Additionally, they determined probability $p$ for each state transition, where a dispatch occurs for the ambulance $i$ nearest to the patient's location. Equation 44 expresses this where $I$ is the set of ambulances, $J$ is the set of demand points, $\lambda_{j}(t)$ is the specific amount of calls which happen at $j$ per unit of time, $\mu$ represents the average service time of a vehicle, and $S_{ki}$ is a state transition at time $k$ for ambulance $i$. The simulation held when analyzed against actual data and was used to solve a MECP. They found that the objective function, in this case, was over-estimating the coverage.

\begin{equation}
    p = \mu \sum_{i \in I}S_{ki} / (\mu \sum_{i \in I}S_{ki}+\sum_{j \in J} \lambda_j(t))
\end{equation}

Lee et al. addressed a hybrid air and ground ambulance problem related to maximizing service coverage by simultaneously locating trauma centers and helicopter ambulances \cite{Lee:2012:SIM:2429759.2429871}. They confronted the issue associated with estimating the busy fraction of available helicopters, which was required to develop an accurate model and maximize the objective function of successfully transporting patients. There modified busy fraction calculation is denoted by Equation 45 where $H_{ij}$ is the set of heliport bases, $L_{h}$ is the service time operated by all helicopters at a particular base, and $K$ is the set of helicopters. A constant of 24 hours was assumed for helicopter operations. Additionally, they considered a location problem where the goal was to maximize the expected number of patients that could be transported within a 60-minute window from the incident’s occurrence. In this scenario, they utilized a combination of DES and integer programming to iteratively update the busy fraction; improving the solution compared to other methodologies of the same type.

\begin{equation}
    p_{ij,k} = \frac{\sum_{h \in H_{ij}}L_{h}}{24k} \;\;\forall \; (i,j)\;\; H_{i,j}\neq \emptyset
\end{equation}

\begin{figure}[ht]
  \centering
  \includegraphics[width=0.6\linewidth]{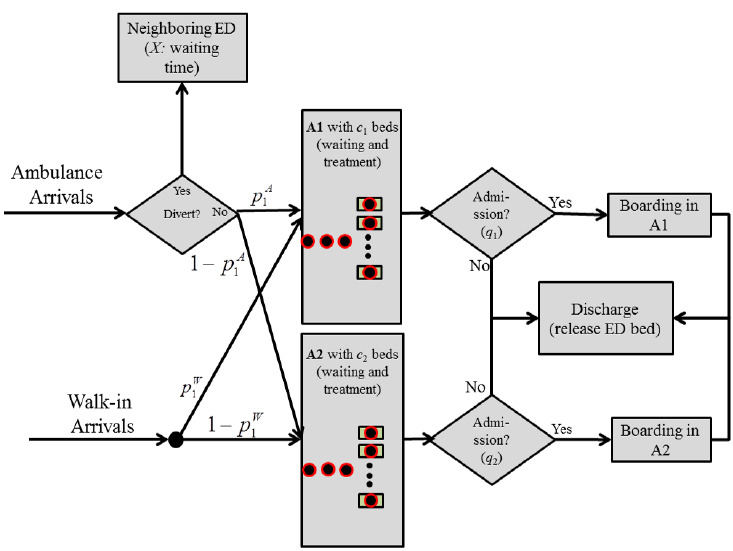}
  \caption{Full representation of patient flow and diversion.}
  \label{3_ramirez}
\end{figure}

"Ambulance diversion" is a method of relieving congestion whereby ambulances bypass a location for another. Ramirez-Nafarrate et al. investigated the effect caused by this diversion strategy, analyzing average waiting times \cite{Ramirez-Nafarrate:2012:CAD:2429759.2429872}. A description of patient flow can be seen in Figure \ref{3_ramirez}, which is essentially separated by how a patient arrives. For their simulation they assumed a "non-homogeneous Poisson process" for arrivals and separated patients based on severity. Different policies were analyzed and included: diversion when all the beds were occupied, policies obtained utilizing Markov decision process formulation, and not allowing diversion at all. The MDP formulation strategy greatly improved the average waiting times when compared to non-diversion and simple policies. Capacity is still something requiring further work as it effected both the fraction of time spent on diversion and the average waiting time. Essentially, small changes in capacity significantly impacted the performance. They suggest that with this consideration one could determine the optimal capacity for a limited budget.

Dibene et al. Modeled the problem to determine if the current placement of emergency services was optimal in Tijuana Mexico. They relied on past data of emergency calls and then resolved a modified version of the DSM using integer linear programming \cite{DIBENE2017107}. The model considered potential base locations, call demand and priority, demand scenarios, demand location, and average travel times. To more closely relate to real world policy, they based their variables off the United States EMS Act. The small radius time standard $r_1$ is set to 10 minutes, the large radius time standard $r_2$ is set to 14 minutes, $\alpha$ which is the fraction of demand covered by within $r_1$  is 0.95, the number of ambulances available $p$ is 11, and the number of $p$ ambulances per base $j$ is 2. They determined that the services should be moved to cover the demand and improve coverage.

In this research, a study was performed utilizing past population and incident data to determine the best position of air ambulance services in the country of Norway. R{\o}islien et al. then formulated and solved an MCLP, whereby the number and location of bases were explored with a constrained time threshold \cite{Rislien2018ComparingPA}. The model determined the optimal allocation of facilities to maximize the number of inhabitants and incidents that could be covered within a certain time window. The experiment was resolved with CPLEX with results suggesting that population density alone was not an appropriate option for determining placement and instead must focus on incident areas. A summary of the results are seen in Table \ref{green_field}, where "greenfield analysis" (assuming no existing bases) and small base adjustment scenarios were compared.

\begin{table}[!ht]
\caption{Coverage comparison utilizing either population density or municipal incident data for existing base structure and a greenfield scenario \cite{Rislien2018ComparingPA}.}
\label{green_field}
\resizebox{\columnwidth}{!}{\begin{tabular}{|l|l|l|l|l|l|l|l|}
\hline
\textbf{Data} & \textbf{Time Threshold} & \textbf{Bases} & \textbf{Population Coverage} & \textbf{Incident Coverage} & \textbf{Base Scenario} & \textbf{Population Coverage} & \textbf{Incident Coverage} \\ \hline
Population    & 45                      & 5                        & 93.32                        & 78.31                      & Use existing           & 96.90                        & 91.86                      \\ \hline
Population    & 45                      & 6                        & 96.29                        & 83.55                      & Relocate one base      & 98.40                        & 93.44                      \\ \hline
Population    & 45                      & 10                       & 100.00                       & 100.00                     & Add one base           & 98.40                        & 93.44                      \\ \hline
Incidents     & 45                      & 7                        & 96.18                        & 94.73                      & Use existing           & 96.90                        & 91.86                      \\ \hline
Incidents     & 45                      & 8                        & 98.22                        & 97.91                      & Relocate one base      & 97.90                        & 96.35                      \\ \hline
Incidents     & 45                      & 10                       & 100.00                       & 100.00                     & Add one base           & 97.90                        & 96.35                      \\ \hline
Population    & 30                      & 9                        & 91.97                        & 69.30                      & Use existing           & 84.70                        & 72.13                      \\ \hline
Population    & 30                      & 12                       & 96.36                        & 81.88                      & Relocate one base      & 87.93                        & 71.50                      \\ \hline
Population    & 30                      & 22                       & 100.00                       & 100.00                     & Add one base           & 88.98                        & 75.35                      \\ \hline
Incidents     & 30                      & 14                       & 93.81                        & 92.04                      & Use existing           & 84.70                        & 72.13                      \\ \hline
Incidents     & 30                      & 16                       & 94.76                        & 96.27                      & Relocate one base      & 86.00                        & 74.77                      \\ \hline
Incidents     & 30                      & 22                       & 100.00                       & 100.00                     & Add one base           & 85.68                        & 76.62                      \\ \hline
\end{tabular}}
\end{table}

Van Den Berg et al. made an argument that to ensure rapid responses, the location of bases and distribution of vehicles among them must be optimized \cite{10.1371/journal.pone.0215385}. They modeled and solved an MECP for urban and rural regions in Norway using a fixed number of bases and variable ambulances. Analysis was performed in a greenfield scenario and also utilized the current implemented base structure. The busy fraction for the problem was varied between four potential possibilities including 0 (single coverage), 0.15 (night), 0.35 (evening), and 0.50 (day). They determined that four out of the five bases were already within an optimal location, although coverage could still be improved up to 97.51\% from the current 93.46\% (see Table \ref{green_field_2}).

\begin{table}[!ht]
\caption{Coverage for differing base values for the current structure and greenfield scenario.}
\label{green_field_2}
\resizebox{0.6\columnwidth}{!}{\begin{tabular}{|l|l|l|}
\hline
\textbf{Number of Bases} & \textbf{Current Coverage (\%)} & \textbf{Greenfield coverage (\%)} \\ \hline
1                        & 45.14                          & 46.96                             \\ \hline
2                        & 68.99                          & 78.24                             \\ \hline
3                        & 82.14                          & 90.26                             \\ \hline
4                        & 91.89                          & 95.84                             \\ \hline
5                        & 93.46                          & 97.51                             \\ \hline
6                        &                                & 98.36                             \\ \hline
7                        &                                & 98.55                             \\ \hline
15                       &                                & 99.40                             \\ \hline
\end{tabular}}
\end{table}

Lanzarone et al. analyzed a two-fold problem involving both ambulance locations and dispatching \cite{Lanzarone:2018:ROA:3320516.3320818}. An ordered list of ambulances was constructed for dispatching to each zone and if there was not an availability from the list then the closest one available was dispatched. They utilized four "recursive optimization-simulation approaches" (ROSA), based on the DES to determine the probability that an ambulance would be busy upon a call. In this scenario a vehicle was either idle or busy, being able to be assigned to a call only if it was idle. Similarly, to the ERTM, this determination was made to better minimize the response time in calls by being able to accurately dispatch to the scene. A comparison between the models determined that more sophisticated models offered a more significant result. The modelling of this problem utilized a heavily modified objective function (see Equation 46) based on the ERTM. It considers those not included in the dispatching list and penalizes $T_p$ if no vehicles are available. As such, an extended list of length $|K|$ was provided for each zone $i$. It continues after the base list, order from closest to farthest, and defined as $\widetilde{y}^{z}_{ij}$. The remaining portion of the problem $(1-q)q^{z-1}$ accounts for the the probability of a vehicle in point $z$ responding to the call.

\begin{equation}
    \sum_{i \in I} \sum_{z \in Z} \sum_{j \in J}(1-q)q^{z-1}d_{i}t_{ji}\widetilde{y}^{z}_{ij} + \sum_{i \in I}d_{i}q^{|K|}T_{p}
\end{equation}

Ingolfsson et al. described an ALP optimization model, minimizing the number of vehicles required and maximizing coverage \cite{optimal_loc}. They considered different service levels, where a level was measured as the fraction of incidents reached within a particular time frame; considering uncertainty by calculating response time as a combination of a random delay summed with random travel time. It was thought that by modeling this randomness a more realistic model could be developed. This design was able to be solved for large scale cities by general-purpose optimizers. The expected coverage was assessed through an approximate hypercube model and optimally solved with the "Premium Solver Platform". The conclusion of this research determined that the inclusion of this type of variability impacted the model significantly and must be considered in actual policy.

Boujemaa et al. took the approach of using stochastic programming and considered cost minimization for transportation, allocation, and even station development \cite{6761400}. A "Sample Average Approximation" method by integer linear program was first used to approximate the model and then a Monte Carlo optimization method called Sample Average Approximation solved random instances of ambulance location and allocation. The results showed a slightly higher cost for the stochastic model compared to deterministic. The stochastic model could aid in avoiding the exposure of patients to greater risk from not satisfying the constraints related to capacity seen in the deterministic variation.

Tavakkoli-Moghaddam et al. considered a pre-disaster situation with the idea of locating temporary emergency stations and ambulance routes \cite{8370579}. The disaster phases are divided based on different points of time; where this study specifically emphasized preparedness. The model was a "bi-objective" variation where the optimal quantity and costs were determined in conjunction with a minimized total response time. Patients were grouped according to priority and then optimal relief stations and routes were calculated. Two objective functions were developed with the first, expressed by Equation 47, minimizing the service completion time among "red code" (serious injuries) and "green code" (slight injuries) patients. Equation 48 describes the second objective as minimizing the setup cost for temporary emergency stations. $w_{r}$ and $w_{g}$ are respectively the priority given to green and red code patients, while $E_{g}$ and $E_{r}$ are the latest service completion times. $s_{h}$ is the cost to establish a hospital in the $h$ region and $u_{h}$ is a binary variable for actually establishing a facility. A study was performed, considering a disaster and the catastrophic results upon densely populated regions. Their model was validated with a $\epsilon$-constraint method, whereby one the objective functions was optimized by using the other as a constraint.

\begin{equation}
    \min w_{g}E_{g} + w_{r}E_{r}
\end{equation}

\begin{equation}
    \min \sum_{h} s_{h}u_{h}
\end{equation}


Zhang and Zeng considered location and possible relocation of ambulances using a robust optimization model with two stages combined with mixed-integer programming for capturing the sequential decision-making process \cite{8574936}. For this, an approximation extension was designed based on the "column-and-constraint" (C\&CG) generation algorithm, producing an optimal or near-optimal solution when used with real-world data. In this research, they additionally considered the uncertainty associated with unavailability and relocation.

\subsection{Ant Colony Optimization (ACO)}
\label{aco_op}
\subsubsection{ACO Description}
"Ant colony optimization" (ACO) is biological metaheuristic technique, which attempts to solve complex problems through simulating the decision process performed by ants as they forage for food \cite{BELL200441, catay2009}. The general process of this is summarized in Algorithm \ref{aco_algorithm}. It begins with lines 1 to 4, where there is a random initialization and the artificial ants move around randomly searching a certain space. Per line 5, once they discover a source they will leave a pheromone trail indicating a path to the food. Lines 7 and 8 indicate that there are good and bad solutions, which increase or decrease the amount of trails. As other artificial ants discover the trails there will be a probability, they will follow the trail themselves. If they decide to take the path, they will leave their own trail and over time create shorter, stronger paths (an optimal solution). There are issues with ACO including an uncertainty in convergence time, the dependency on randomness, and the difficulty structuring certain problems \cite{HOOS200561}. Regardless, this type of algorithm is excellent for techniques such as routing and location determination, where a vehicle or facility takes the place of the simulated ant. 

\medskip
\begin{algorithm}[H]
\label{aco_algorithm}
 Initialize pheromone trails for $m$ ants\;
 \While{stop condition not satisfied}{
  \For{each $m$ artificial ant} {
    Construct a solution\;
    Update local pheromone trail\;
  }
  Update best solution\;
  Perform pheromone update;\
 }
 \caption{Ant Colony Optimization Pseudo-code}
 \label{aco_code}
\end{algorithm}
\medskip

\subsubsection{ACO Solutions}
Benabdouallah et al. investigated a dynamic coverage problem for the Casablanca region of Morocco to minimize fitness (total lateness) \cite{7831330}. The fitness was calculated using DES as seen in Equation 49. In their model $k$ was the intervention, $K$ was the set of demands, $t_{k}$ was the date of demand arrival, and $d_{k}$ was the patient's arrival to the hospital by the ambulance. ACO was used in conjunction with a local search guide, whose purpose was to find an improved fitness at the current generated solution. The results were compared against a heuristic method, where ACO was the better of the two in terms of fitness (see Table \ref{aco_tab_results}).

\begin{equation}
    \min \sum_{k\in K}d_{k} - t_{k}
\end{equation}

\begin{table}[!ht]
\caption{Results upon randomized data and compared again a heuristic.}
\label{aco_tab_results}
\resizebox{0.55\columnwidth}{!}{\begin{tabular}{|l|l|l|l|l|}
\hline
\textbf{Instances}                  & 20    & 22    & 27    & 32    \\ \hline
\textbf{Hospitals}                  & 4     & 6     & 4     & 5     \\ \hline
\textbf{Heuristic Fitness (min)}    & 4,394 & 3,798 & 4,699 & 6,598 \\ \hline
\textbf{Heuristic Solving Time (s)} & 114   & 184   & 174   & 554   \\ \hline
\textbf{ACO Fitness (min)}          & 4,381 & 3,783 & 4,686 & 6,584 \\ \hline
\textbf{ACO Solving Time (s)}       & 5,312 & 5,665 & 7,094 & 7,581 \\ \hline
\end{tabular}}
\end{table}

\begin{figure}[h]
  \centering
  \includegraphics[width=0.6\linewidth]{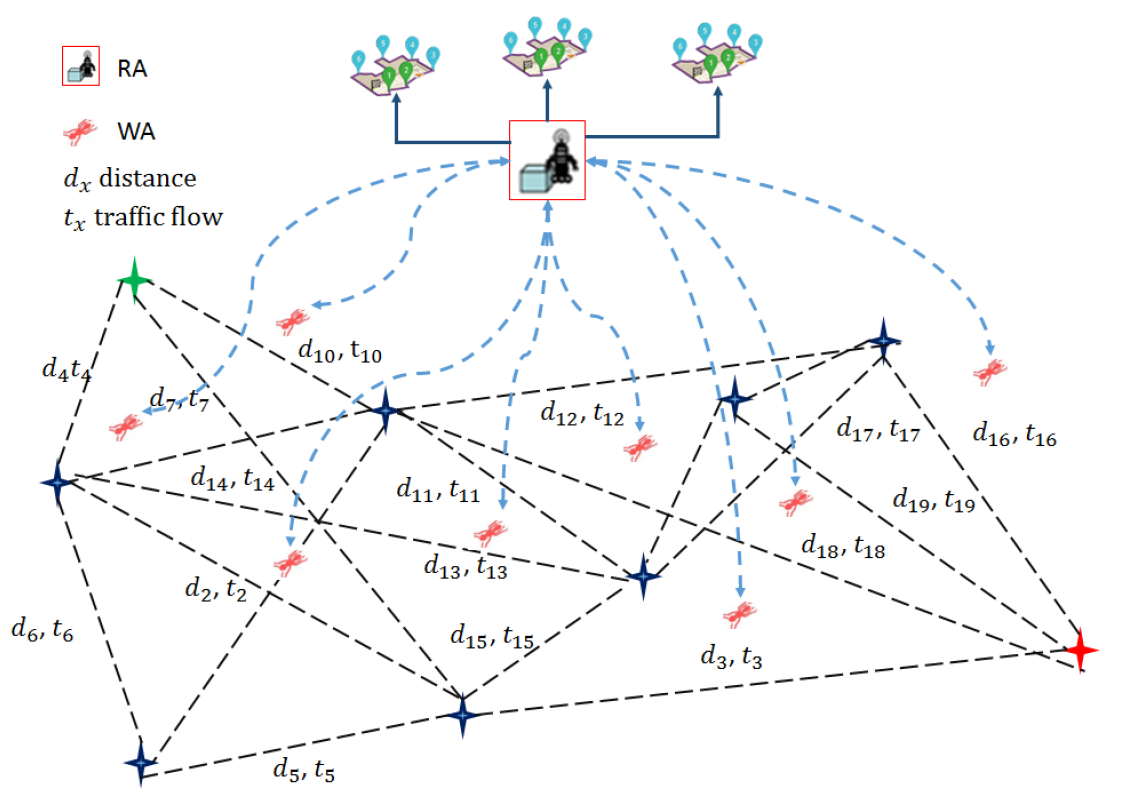}
  \caption{ACO model for finding a route to an ambulance.}
  \label{5_mouchine}
\end{figure}

Mouhcine et al. proposed a distributed solution using ACO to determine the optimal paths for emergency vehicles \cite{8370582}. The solution sought to minimize the travel time while constrained by obstacles such as traffic, speed limit, available vehicles, and facility positioning. The system generated a dynamic path, determining the shortest paths based on a set of intelligent agents. As seen in Figure \ref{5_mouchine}, the routing age (RA) sends out worker agents (WA) to determine optimal ambulance routes. Following each iteration, another path is determined and best ones are selected. Mouhcine et al. suggested that further research needs to be done on the solution, as convergence to near-optimal is not guaranteed.

Javidaneh et al. attempted to resolve an ambulance routing problem by implementing an approach using ACO \cite{ant_rout_am}. The problem assumed that the number and location of facilities, vehicles, and patients were known; as well as the capacity of both facilities and vehicles. The goal of the solution was to route ambulances in the least amount of time possible, while constrained by route lengths and hospital capacities. For this solution the algorithm inputted an adjacency matrix representation where the graph displayed the location of people requiring assistance, the capacity of the hospitals and ambulances, and the number of people at each facility. The output for this model was simply the routes for each respective ambulance satisfying the objective function. Results of the algorithm were comparable to real-world, displaying the usefulness of ACO in this type of optimization problem.

\subsection{Genetic Algorithm (GA)}
\label{ga_op}
\subsubsection{GA Description}
"Genetic algorithms" (GA) are part of a group of stochastic optimization techniques, where a function is minimized or maximized with randomness introduced into the process \cite{BAKER2003787}. It is based on Darwin’s theory of evolution and essentially simulates the idea of natural selection, where it is repeated until hopefully a near-optimal solution is found. Though the steps may be altered in general they are selection, crossover, and mutation. As seen in Algorithm \ref{algorithm_ga}, it starts with a random initialized population and then assesses the fitness (objective function value) to choose points (called “parents”) contributing to the next evolution (lines 1 to 3). Two parents will combine, and the new “children” form the next generation in the system (line 5). Lastly, random changes known as mutations will be added to the population to add diversity to preceding generations and stop the optimization from ending up in a local minimum (line 6). At the end of a generation the fitness is evaluated and the steps are repeated (line 7). A limitation with a GA is the fact that design choices must be carefully made, as certain choices may lead to unfavorable results (a local optima) or even an incorrect answer \cite{LEARDI2009631}. Additionally, GA’s are subject to the initial population, making later convergence a possible issue.

\medskip
\begin{algorithm}[H]
\label{algorithm_ga}
 Generate a starting population\;
 Asses the population based on the fitness\;
 \While{stop condition not satisfied}{
    Perform the Selection operation\;
    Perform the Crossover operation\;
    Perform the Mutation operation\;
    Evaluate the new population through fitness\;
 }
 \caption{Genetic Algorithm Pseudo-code}
 \label{genetic_code}
\end{algorithm}
\medskip

\subsubsection{GA Solutions}
Iannoni et al. embedded a spatially distributed queuing (hypercube) model into hybridized GA’s to optimize the operations of ambulances \cite{IANNONI2009528}. The method considered the location of ambulance facilities along a highway, as well as how to district the responses. The model was adapted to analyze either a "single dispatch" (model 1) or "single and double dispatch" (model 2). Essentially, the model could either optimize the coverage areas to reduce response times or resolve workload imbalances. It was validated with DES and found in real scenarios that policies could be improved by facility relocation. Step 1 was to apply optimize the location of ambulance bases through a location GA. Following this, step 2 started from the solution generated by step 1 and improved it through the modified districting GA/hypercube model. A summary of the results are in Table \ref{Iannoni_tab_results} where $\overline{\rm T}(x)$ is the primary minimization goal as the mean region wide travel time, $\sigma_{p}(x)$ is the imbalance of ambulance workloads expressed as standard deviation, and $P_{t>10}(x)$ is the fraction of calls not serviced in 10 minutes.


\begin{table}[!h]
\caption{Results of GA/hypercube on minimizing travel time.}
\label{Iannoni_tab_results}
\resizebox{0.8\columnwidth}{!}{
\begin{tabular}{|l|l|l|l|l|l|}
\hline
\textbf{Ambulances} & \textbf{Measure}      & \textbf{Step 1 Improvement (\%)} & \textbf{Runtime (h)}    & \textbf{Step 2 Improvement(\%)} & \textbf{Runtime (h)}    \\ \hline
\multirow{3}{*}{6}  & $\overline{\rm T}(x)$ & 19.97                            & \multirow{3}{*}{0.0038} & 0.47                            & \multirow{3}{*}{0.0034} \\ \cline{2-3} \cline{5-5}
                    & $\sigma_{p}(x)$       & 46.86                            &                         & 4.16                            &                         \\ \cline{2-3} \cline{5-5}
                    & $P_{t>10}(x)$         & 39.12                            &                         & -0.29                           &                         \\ \hline
\multirow{3}{*}{8}  & $\overline{\rm T}(x)$ & 12.49                            & \multirow{3}{*}{0.0828} & 0.48                            & \multirow{3}{*}{0.156}  \\ \cline{2-3} \cline{5-5}
                    & $\sigma_{p}(x)$       & 33.14                            &                         & -14.41                          &                         \\ \cline{2-3} \cline{5-5}
                    & $P_{t>10}(x)$         & 29.36                            &                         & -1.36                           &                         \\ \hline
\multirow{3}{*}{10} & $\overline{\rm T}(x)$ & 16.90                            & \multirow{3}{*}{1.518}  & 0.33                            & \multirow{3}{*}{2.120}  \\ \cline{2-3} \cline{5-5}
                    & $\sigma_{p}(x)$       & 59.65                            &                         & -5.28                           &                         \\ \cline{2-3} \cline{5-5}
                    & $P_{t>10}(x)$         & 31.96                            &                         & -1.41                           &                         \\ \hline
\end{tabular}
}
\end{table}

\bigskip

Chuang and Lin confronted an ALP based on MECP and DSM \cite{doi:10.1080/10170660709509061}. They proposed a combination model known as the "maximum expected covering location problem with double standard" (MEXCLP-DS) to solve in a probabilistic situation and meet the coverage not provided by the previous individually. A formulation of this combination is represented in Equations 50 through 53, and uses a similar set of variables to the MECP and DSM discussed in Section 2. Different varialbes include $h_{k}$ as the probability of a call occurring at node $j$ and $l$ as the total number of ambulances. The objective function shown in Equation 50 maximizes the expected number of demands that can be covered. Equation 51 ensures that a proportion of the demand is covered based on $\alpha$, Equation 52 states that the total amount of vehicles at every node must be larger than the amount of facilities covering a node, and Equation 53 prevents the number of vehicles at every node from being greater than the total number of vehicles. The models were resolved with a GA using a combination of directed and stochastic searches, resulting in the demand being able to be 100\% covered with an 8-minute standard arriving time.

\begin{align}
    & \text{max}
    & &\sum_{i = 1}^{l}\sum_{j = 1}^{m}d_{i}(1-q)q^{l-1}y_{ij}h_{j} \\
    & \text{subject to} 
    & & \sum_{i = 1}^{l}\sum_{j = 1}^{m}d_{i}y_{ij}h_{j} \:\geq \alpha \sum_{i = 1}^{l}\sum_{j = 1}^{m}d_{i}h_{j} \\
    &&& \sum_{j = 1}^{m}x_{j} \:\geq \sum_{i}^{l}\sum_{j}^{m}y_{ij} \\
    &&& \sum_{j = 1}^{m}x_{j} \:\leq l
\end{align}

Tlili et al. discussed an ARP for emergency medical services and utilized real-world data \cite{TLILI20181585}. They modeled the problem in what they referred to as a simple and open ARP, where the goal was to serve a greater number of patients, while utilizing the same amount resources and minimizing the total travel distance. In terms of formulation the problem used a traditional VRP-based design; however, also considered a total budget constraint. This is shown in Equation 54, stipulating that the travel cost cannot exceed to total budget $C$ for all ambulances $p$. To resolve this a GA was implemented; however, utilized a customized recombination operator. In small instances, the solution was solved efficiently.

\begin{equation}
    \sum_{i\in I}\sum_{j\in J}x_{ijk}c_{ij} \:\leq C_{max}^k, \forall \: 1 \leq k \: \leq p 
\end{equation}

Sasaki et al. made an argument that with changing demographics, ambulance response times are decreasing steadily \cite{Sasaki_10.1186/1476-072X-9-4}. They used predictive techniques to determine future EMS cases and then compared current and future optimal locations. The model was resolved with a modified grouping GA to determine potential sets of ambulance locations. Without a GA, a calculated brute force method would have needed to consider $4.415e + 35684$ permutations. The results concluded that response times could be decreased by about a minute if the updated locations are considered.

Fogue et al. concentrated on non-emergency patient transport services for patients that did not require urgent care yet still required hospital transportation \cite{Fogue:2016:NPT:2972705.2972878}. For this problem, a human operator usually developed a suboptimal solution, which they aimed to overcome with a novel algorithm referred to as the non-urgent transport routing algorithm (NURA). This method utilized a GA as part of its core to explore the solution space and generate detailed routes for ambulance services. A scheduling algorithm was then generated the specific plan for each mission. The results of the algorithm appeared to outperform human experts in similar conditions and reduced the average waiting time (see Table \ref{NURA_tab_results}).


\begin{table}[!ht]
\caption{Comparison performance between human experts and NURA.}
\label{NURA_tab_results}
\resizebox{0.55\columnwidth}{!}{\begin{tabular}{l|l|l|l|l|}
\cline{2-5}
                                        & \multicolumn{2}{l|}{\textbf{Patient Waiting Time (min)}} & \multicolumn{2}{l|}{\textbf{Ambulance Usage}} \\ \hline
\multicolumn{1}{|l|}{\textbf{Scenario}} & \textbf{Human}              & \textbf{NURA}              & \textbf{Human}         & \textbf{NURA}        \\ \hline
\multicolumn{1}{|l|}{\textbf{1}}        & 42                          & 36.40                      & 70.98\%                & 77.59\%              \\ \hline
\multicolumn{1}{|l|}{\textbf{2}}        & 84                          & 78.00                      & 64.29\%                & 83.16\%              \\ \hline
\multicolumn{1}{|l|}{\textbf{3}}        & 67                          & 60.67                      & 66.14\%                & 81.44\%              \\ \hline
\end{tabular}}
\end{table}

Pacheco et al. addressed a static ALP for a fleet in Tijuana, Mexico \cite{7257213}. In the research, they modified the model as a multi-source Weber problem to ensure it was designed as continuous and included EMS time thresholds. In this problem the aim is to locate $m$ facilities and allocate the demand to those chosen. The objective, as shown in Equation 55, was to reduce the sum of the weighted distances from facilities to demand point. $K$ is the set of demand points, $w_k$ is the associated weight value, $d_k$ is the distance from $k$ to a facility, $(x,y)$ are location coordinates, $p_{k}$ is a particular point where an emergency occurs, and $v_i$ is a binary variable allowing for a selection between the travel time of $d(p_{i}, (x,y))$ and the time tolerance threshold. For the solution, a GA was implemented and compared against the current placement; results indicating improvement over existing selections for expected travel time $t_{k}$ between two locations (see Table \ref{Pacheco_tab_results}).

\begin{align}
    & \text{max}
    & &\sum_{k \in K}\sum_{j = 1}^{m}w_{k}d(p_{k}, (x_{j}, y_{j}))v_{kj} \\
    & \text{subject to} 
    & & \sum_{j = 1}^{m}v_{kj} = 1 , \forall \: k \\
    &&& v_{kj} \in \{0,1\} , \forall \: k, j = 1,...,m \\
    &&& x,y \in \mathbb{R}^m
\end{align}

\begin{table}[!ht]
\caption{GA performance upon different demand points.}
\label{Pacheco_tab_results}
\resizebox{0.6\columnwidth}{!}{\begin{tabular}{|l|l|l|l|l|l|}
\hline
\textbf{Demand Points}       & 50     & 75     & 100    & 125     & 150    \\ \hline
\textbf{Average Time (min)}  & 5.76   & 8.64   & 11.51  & 14.49   & 16.90  \\ \hline
\textbf{SD Time (min)}       & 0.0960 & 0.1249 & 0.0182 & 0.01810 & 0.2667 \\ \hline
\textbf{Average Solution $t_{k}$} & 5.3745 & 6.4391 & 5.9346 & 5.9683  & 5.3152 \\ \hline
\textbf{SD Solution $t_{k}$}      & 2.2183 & 0.8695 & 1.5494 & 1.7669  & 1.7736 \\ \hline
\end{tabular}}
\end{table}

Benabdouallah and Bojji modeled the problem as a dynamic double standard model (DDSM) to garner the best distribution of ambulances to bases \cite{Benabdouallah:2017:CGA:3128128.3128136}. They then compared the resulting coverage optimized by a GA and ACO approach. The goal of these models was to minimize the lateness of ambulances to disaster locations and were based on random instances within a two-period day. The GA approach garnered a better-minimized fitness when compared to the exact solution extracted by CPLEX, with the latter not being able to solve beyond a certain threshold (see Table \ref{Benabdouallah_tab_results}).

\begin{table}[!ht]
\caption{Comparison between GA and CPLEX experiments.}
\label{Benabdouallah_tab_results}
\resizebox{0.8\columnwidth}{!}{\begin{tabular}{|l|l|l|l|l|}
\hline
\textbf{Sectors} & \textbf{GA Fitness (min)} & \textbf{CPLEX Fitness (min)} & \textbf{GA CPU(s)} & \textbf{CPLEX CPU(s)} \\ \hline
2                & 643.8046666666667         & 643.8046666666111            & 0.005              & 0.42                  \\ \hline
3                & 685.8046666666667         & 685.8046666665872            & 0.005              & 0.27                  \\ \hline
4                & 753.0294666666666         & 745.8046666665633            & 0.004              & 6.49                  \\ \hline
5                & 830.5627333333333         & *                            & 0.012              & *                     \\ \hline
\end{tabular}}
\end{table}

Ramirez-Nafarrate et al. proposed a simulation to explore ambulance diversion (AD), where an overcrowded emergency department requested an ambulance to bypass their location \cite{Nafarrate:2011:DCA:2431518.2431667}. Their research was based on DES with the goal of analyzing the effect of AD on patient flow within the care system. The simulation was utilized to determine the parameters required for the diversion policies and reduce the expected travel time. Additionally, the research proposed a GA for designing and implementing new policies, the results of which displayed significant improvement over current patient flow methods.


\subsection{Local Search-Based Solutions}
\label{local_op}
\subsubsection{Local Search Description}
"Local search" forms the basis for several optimization techniques and in its simplest form is itself a process for solving computationally intensive problems. A local search-based algorithm moves between multiple possible solutions within a search space, making small localized changes as it progresses. It will only stop once it reaches an appropriate near-optimal solution or another specific stopping condition (such as time) occurs. A key issue with local search in its basic form is that it commonly only attains a local optima \cite{Funke2005}. As such, there are multiple algorithms which have been adapted and extended this solution.

\subsection{Neighbourhood Search Description}
Variations of local search manipulate a set of data within a certain space known as a neighbourhood. Three common neighbourhood-based techniques are "large neighborhood search" (LNS), "variable neighborhood search" (VNS), and "nearest neighbor search" (NNS). LNS gradually improves an initial solution by modifying an incumbent answer over an exceptionally large area \cite{lns_book}. The purpose of this is to minimize the potential for getting locked into local minimums as the improvements are much greater. VNS updates a current solution by exploring distant neighbourhoods and adjusting if there is a more optimal value \cite{Hansen2010}. In this case multiple neighbourhoods are explored and will repeat until an optimum is discovered. NNS is not just one, but a group of algorithms optimizing by determining points within a certain set that are more similar or can be adjusted based on another defined criterion \cite{nns_article}. 

\subsubsection{Tabu Search Description}
"Tabu Search" further extends the idea of the neighbourhood-based algorithms by excluding recently explored areas within a search space and possibly allowing moves that would not improve the objective. A list of previously visited answers are held within a Tabu list, whose goal is to prevent a search from becoming locked into a local optimum. In general it only records recent moves and will not allow a solution which has been explored within a particular period. This can optionally be ignored if it would otherwise improve the fitness. The list usually clears certain answers from it after a period of time, although the size of the list can vary depending on the problem \cite{EDELKAMP2012633}. The pseudo-code of the Tabu search steps are shown in Algorithm \ref{algorithm_tabu}. Each solution is represented by a state, where a state is the organization of variables into a specific solution. From lines 1 to 3, an initial solution state is generated and saved to the initialized empty Tabu list. Then from lines 4 to 10 new neighbourhood solutions are explored and the solutions are updated if they have not been added to the current list. Once a new improved state is discovered it is placed inside of the Tabu list at line 8 and the cycle continues until the user defined stopping condition is satisfied.

\medskip
\begin{algorithm}[H]
\label{algorithm_tabu}
 Calculate initial best state\;
 Initialize the tabu list\;
 Save initial state to tabu list\;
 \While{stop condition not satisfied}{
 \For{each candidate $n$ in a neighbourhood} {
    \If{$n$ is not in the tabu list and improves the fitness} {
        Update current states\;
        Add new state to the tabu list\; 
    }
 }
 }
 \caption{Tabu Search Pseudo-code}
 \label{genetic_code}
\end{algorithm}
\medskip

\subsubsection{Local Search Solutions}
In this research, Talarico et al. considered an ARP where a disaster occurred and a large group required simultaneous aid \cite{TALARICO2015120}. For the scenario, the patients were grouped into those who were slightly injured and those who were seriously injured. Two models were proposed to obtain a plan that would minimize the time; a "3-index" and "2-index" model. As disaster response requires immediate action, they also introduced a LNS metaheuristic to solve the solution in a much more efficient time frame. An initial solution is generated by either an insertion or construction heuristic, then LNS improves the solution with nine different local search operators, and lastly the current solution is partially destroyed then reconstructed by a repair operator to reach unexplored areas within the space. The algorithm approached near-optimal, as summarized by Table \ref{Talarico_tab_results}, and was additionally able to consider the priority of patients.

\begin{table}[!ht]
\caption{Results achieved by the 3-index and 2-index models.}
\label{Talarico_tab_results}
\resizebox{0.6\columnwidth}{!}{\begin{tabular}{c|l|l|l|l|}
\cline{2-5}
\multicolumn{1}{l|}{}                                         & \textbf{Number of Patients}           & 10   & 25   & 50   \\ \hline
\multicolumn{1}{|c|}{\multirow{5}{*}{\textbf{3-index Model}}} & \textbf{Number of Feasible Solutions} & 108  & 107  & 82   \\ \cline{2-5} 
\multicolumn{1}{|c|}{}                                        & \textbf{Number of Optimal Solutions}  & 79   & 9    & 0    \\ \cline{2-5} 
\multicolumn{1}{|c|}{}                                        & \textbf{Lower Bound}                  & 611  & 279  & 256  \\ \cline{2-5} 
\multicolumn{1}{|c|}{}                                        & \textbf{Fitness}                      & 881  & 1399 & 2205 \\ \cline{2-5} 
\multicolumn{1}{|c|}{}                                        & \textbf{CPLEX Time (s)}               & 1416 & 3524 & 3600 \\ \hline
\multicolumn{1}{|c|}{\multirow{5}{*}{\textbf{2-index model}}} & \textbf{Number of Feasible Solutions} & 108  & 108  & 108  \\ \cline{2-5} 
\multicolumn{1}{|c|}{}                                        & \textbf{Number of Optimal Solutions}  & 79   & 21   & 3    \\ \cline{2-5} 
\multicolumn{1}{|c|}{}                                        & \textbf{Lower Bound}                  & 617  & 359  & 270  \\ \cline{2-5} 
\multicolumn{1}{|c|}{}                                        & \textbf{Fitness}                      & 881  & 1069 & 1363 \\ \cline{2-5} 
\multicolumn{1}{|c|}{}                                        & \textbf{CPLEX Time (s)}               & 1165 & 3011 & 3504 \\ \hline
\end{tabular}}
\end{table}

Schmid and Doerner developed a "multi-period dynamic" variation of the ALP, allowing for the repositioning of vehicles and based on the DSM \cite{SCHMID20101293}. The model was structured with MILP and the optimal was solved for simultaneous points, while also considering time-dependent data. A modification to the standard DSM objective function can be seen in Equation 59. For this modification, a DSM has to be simultaneously considered for every instance $t \in T$. The set of demand locations is represented by $V$ and the set of vehicle locations is $W$. Since relocation is possible, $\beta$ is a penalty for performing this task. In terms of achieving near-optimal, Schmid and Doerner resolved the model with VNS; randomizing starting points, performing a shaking phase to explore the solution space, and improving further with local search. On average the algorithm was able to find a comparable solution to the CPLEX optimizer in a drastically decreased time.

\begin{equation}
    \max \sum_{t\in T} \left( \sum_{i\in V}d_{i}x_{i}^{2,t} - \beta \sum_{i,j\in W}r_{ij}^t \right)
\end{equation}

Da\u{g}layan and Karakaya proposed a solution for effectively scheduling ambulances following a disaster and designed it as a "capacitated vehicle routing problem" (CVRP) \cite{karakaya_2016}. The research aimed to minimize the number of routes and reduce average travel time to hospitals. For this task, they developed a GA and evaluated it against an NNS heuristic. The algorithms were compared against three types of scenarios: light damage (1 to ambulance capacity), medium damage (1 to 8), and heavy damage (ambulance capacity to 8). Following simulation-based experiments on common CVRP benchmarks, the results showed that the GA outperformed the NN method, achieving shorter tour distances (see Figure \ref{nns_ga_tab_results}). A potential downside to this method is in the fact that it was limited to only a single facility and ambulance, though they admitted further study needs to be completed regarding this.

\begin{table}[!ht]
\caption{Improvement of the GA over NNS for known CVRP Benchmarks.}
\label{nns_ga_tab_results}
\resizebox{0.8\columnwidth}{!}{\begin{tabular}{|l|l|}
\hline
\textbf{Benchmark} & \textbf{Average Tour Length Improvement of GA over NNS (\%)} \\ \hline
ulysses-n22-k4     & 21.88                                                        \\ \hline
bays-n29-k5        & 12.77                                                        \\ \hline
A-n38-k5           & 9.65                                                         \\ \hline
A-n40-k5           & 9.63                                                         \\ \hline
E-n51-k6           & 9.05                                                         \\ \hline
P-n55-k10          & 8.85                                                         \\ \hline
P-n60-k10          & 7.27                                                         \\ \hline
B-n64-k9           & 6.12                                                         \\ \hline
E-n76-k9           & 6.08                                                         \\ \hline
P-n101-k4          & 5.62                                                         \\ \hline
\end{tabular}}
\end{table}

Gendreau et al. designed and modeled an ALP after the double coverage variation \cite{GENDREAU199775}. The objective was to maximize the coverage using two ambulances, constrained by actual requirements imposed by EMS service laws. Real and randomly generated data points were used in conjunction with a Tabu Search heuristic, approaching near-optimal results in a reasonable computing time compared to the CPLEX optimizer. While the algorithmic solution approaches near-optimal, the calculation time is only marginally better than CPLEX. The computational resources available could have been a factor as this paper is now older and would most likely have a better solution if utilized on a modern CPU.

Erdogan et al. developed two models for maximizing the coverage by scheduling ambulance crews \cite{sched_max_cov}. The first sought to maximize the aggregate expected coverage which was the ratio between the expected number of calls covered and the total number of calls. This is represented in Equation 60 where $\delta_{ij}$ is number of expected calls covered at hour $i$ by adding vehicle $j$, $y_{ij}$ is a binary variable for setting whether the total amount of ambulance crews during the first hour $i$ is at minimum $j$, $e_i$ is the average number of calls for hour $i$, and $h$ is number of hours in the planning horizon. 
The second was a "lexicographic biobjective" model (Equation 61) where the goal was to maximize both the minimum expected coverage over every preceding hour and the aggregate expected coverage. In this model, $w$ is equal to the minimum expected coverage over every hour. The models were uniquely optimized with a parallel variation of the Tabu Search, outperforming similar previous approaches.

\begin{equation}
    \max \frac{\sum_{i=1}^h \sum_{j=1} \delta_{ij}y_{ij}}{\sum_{i=1}^h e_i}
\end{equation}

\begin{equation}
    \max \left(w,\frac{\sum_{i=1}^h \sum_{j=1} \delta_{ij}y_{ij}}{\sum_{i=1}^h e_i}\right)
\end{equation}

Oberscheider and Hirsch aimed to ensure efficient transport for non-emergency patients utilizing real-patient data from lower Austria's Red Cross \cite{10.1186/s12913-016-1727-5}. The work claimed it contrasted with prior variations as they considered non-static service times that depended on the combination of patients, their transport mode, the vehicle type, and the pickup or delivery locations. Their model is based on the "static multi-depot heterogeneous dial-a-ride problem" (MD-H-DARP) and to solve it they first generated all combinations of patient transports given a set of constraints. A "set partitioning" action was then completed upon the previous generation and an initial solution was generated. They then inputted these combinations into a Tabu Search and further optimized the routing. The strategy was determined to be an improvement over manual scheduling.


Repoussis et al. presented a MILP model to provide operation guidance, routing, and scheduling for mass casualty incidents \cite{REPOUSSIS2016531}. The aim was to allocate minimal resources while reducing the total flow and response times. To resolve this problem, they developed a "hybrid multi-start local search" method which involved a MIP-based construction heuristic followed by an iterated Tabu Search. The algorithm initially uses a "greedy randomized scheme" to find and fix a portion of the assignment and then the reduced problem is optimally solved. These initial solutions form a high-quality upper bounds and are the inputs into the iterated Tab Search algorithm for further improvement. This process is repeated until the condition for termination is met.

\begin{table}[!ht]
\caption{Computation Time for ACO and Tabu Search.}
\label{aco_tabu_results}
\resizebox{0.7\columnwidth}{!}{\begin{tabular}{|l|l|l|}
\hline
\textbf{Region} & \textbf{ACO Computation Time} & \textbf{Tabu Search Computation Time} \\ \hline
Vorarlberg      & 8 s                           & 2 s                                  \\ \hline
Salzburg        & 80 s                          & 15 s                                 \\ \hline
Lower Austria   & 3.5 h                         & 20 m                                 \\ \hline
Carinthia       & 140 s                         & 20 s                                 \\ \hline
Upper Austria   & 1.0 h                         & 6 m                                  \\ \hline
Burgenland      & 18 s                          & 10 s                                 \\ \hline
Styria          & 1.33 h                        & 6 m                                  \\ \hline
The Tyrol       & 6.5 m                         & 2.5 m                                \\ \hline
\end{tabular}}
\end{table}

\subsection{Particle Swarm Optimization (PSO)}
\label{pso_op}

\subsubsection{PSO Description}
"Particle swarm optimization" (PSO) is an algorithm inspired by the social behaviours of insect swarms, fish schools, and flocks of birds. The technique simulates the interaction between members of a set and how they share information between each other. A search in a solution space is performed by a group of agents (the particles or candidate solutions), who are adjusted closer to an optimized answer by stochastic and deterministic parts. The particles are influenced by both a respective local and a global best case, updating the positions accordingly \cite{MARTINEZ201915}. The full process is expressed in Algorithm \ref{algorithm_pso}, starting with an initialization of $m$ particles at line 1. From lines 3 to 8 each particle determines their fitness and can change if a local improvement is discovered. Once each particle has performed these actions, then at line 9 the best global fitness is determined from the local best. Finally, at lines 11 and 12 the particle velocity is determined based on the global best and then each one updates their position among the space.

\medskip
\begin{algorithm}[H]
\label{algorithm_pso}
 Generate and initialize $m$ particles\;
 \While{stop condition not satisfied}{
 \For{each $m$ particle} {
    Determine the current fitness\;
    \If{fitness is better than the current best local} {
        Update the best local fitness\;
    }
 }
 Select particle with the best fitness and set this to the best global fitness\;
 \For{each $m$ particle} {
    Perform the calculation for particle velocity\;
    Perform the update of the particle position\;
 }
 }
 \caption{Particle Swarm Optimization Pseudo-code}
 \label{genetic_code}
\end{algorithm}
\medskip

\subsubsection{PSO Solutions}
Tlili et al. made the argument that transferring patients is a significantly more difficult situation when occurring in a disaster situation as there may be individuals requiring simultaneous care \cite{Tlili:2017:SAS:3141923.3142198}. The problem was modeled using two methods: "open vehicle routing problem" (OVRP) and a "vehicle routing problem with pickup and delivery" (VRPPD). Their optimization solution revolved around an initial stage based on the petal algorithm known as cluster-first route-second. In this method, patients were grouped so that individuals in the same group could be serviced by the same vehicle. Following this, they utilized PSO to attain the best solution while avoiding local minima and defeating a compared GA in the majority of instances (see Table \ref{pso_tab_results}).

\begin{table}[!ht]
\caption{Comparison of PA-PSO and GA.}
\label{pso_tab_results}
\resizebox{0.3\columnwidth}{!}{\begin{tabular}{|l|l|l|}
\hline
\textbf{Instance Label} & \textbf{PA-PSO} & \textbf{GA}    \\ \hline
A-n32-k5                & \textbf{950}    & 957            \\ \hline
A-n33-k5                & \textbf{765}    & 781            \\ \hline
A-n33-k6                & 835             & \textbf{798}   \\ \hline
A-n34-k5                & \textbf{920}    & 923            \\ \hline
A-n36-k5                & \textbf{891}    & 1,019          \\ \hline
A-n37-k5                & \textbf{800}    & 959            \\ \hline
A-n37-k6                & 1,135           & \textbf{1,115} \\ \hline
A-n38-k5                & \textbf{892}    & 974            \\ \hline
A-n39-k5                & \textbf{999}    & 1,095          \\ \hline
A-n39-k6                & \textbf{971}    & 1,257          \\ \hline
A-n45-k6                & \textbf{1,171}  & 1,452          \\ \hline
A-n45-k7                & \textbf{1,385}  & 1,397          \\ \hline
B-n31-k5                & \textbf{696}    & 736            \\ \hline
B-n34-k5                & 936             & \textbf{865}   \\ \hline
B-n35-k5                & \textbf{1,077}  & 1,263          \\ \hline
B-n38-k6                & \textbf{888}    & 925            \\ \hline
B-n39-k5                & \textbf{718}    & 744            \\ \hline
B-n41-k6                & \textbf{882}    & 1,105          \\ \hline
B-n43-k6                & \textbf{874}    & 1,047          \\ \hline
B-n44-k7                & \textbf{1,037}  & 1,126          \\ \hline
P-n20-k2                & \textbf{222}    & \textbf{222}   \\ \hline
P-n19-k2                & \textbf{219}    & 240            \\ \hline
\end{tabular}}
\end{table}

Hatta et al. investigated solving an MCLP, maximizing the coverage of a fixed number of ambulances \cite{psw_mc}. The problem was formulated as a grid where disaster locations could host at most one ambulance and were randomly generated. The model was then resolved with PSO and compared against a random search technique of 300,000 iterations, identifying a globally optimal solution within an acceptable time frame. For comparison, the random search attained only a coverage of 75.44\%, while the PSO achieved 92.52\%. Each solution of the MCLP was expressed as a particle and each one has an $n$ vector representing an ambulance location. The PSO utilized the standard algorithm with Equation 62 for updating the position and Equation 63 for updating the velocity. In these equations $pbest$ is the local best for a particle, $gbest$ is global best based on the fitness evaluation, $w$ is the inertia weight for exploring the space, $r_1$ and $r_2$ are random binary numbers, $w$ and $c$ are weight factors, $v$ is the velocity of a particle, and $s$ is a particle's position.

\begin{equation}
    s_{i}^{k+1} = s_{i}^{k} + v_{i}^{k+1}
\end{equation}

\begin{equation}
    v_{i}^{k+1} = wv_{i}^{k} + c_1 r_1 (pbest_i - s_{i}^k) + c_2 r_2 (gbest - s_{i}^k)
\end{equation}

\bigskip

\subsection{Clustering}
\label{cluster_op}
\subsubsection{Clustering Description}
"Clustering" is a greater technique for a collection of algorithms that seek to group a dataset into separate similar groups (clusters).  There is not one single methodology for clustering and can range from grouping by distance between points or based on the data density. For location problems, centroids can be determined and facilities can be plotted relative to these spots \cite{1427769}. In terms of VRP, clustering can be used to subdivide the problem by clustering locations based on similar features and then designing optimal routes based on the resulting clusters \cite{8128610}. 

\subsubsection{Clustering Solutions}
Kamireddy et al. described an ambulance routing strategy that confronted issues with increased emergency cases due to nonlinear increases in population and multiple requests from the same location \cite{7919560}. As increasing the number of resources can be quite expensive, they opted for improving the optimization of existing ones. They implemented a multi-stage clustering methodology which sought to place ambulances in a particular location, reducing the average distance covered. The thought of the approach was that placing an ambulance near a cluster would decrease the distance to a patient. Different clustering algorithms were utilized including "K-Mean", "Weighted K-Mean", "Density-Based", and "Directly Reachable". The first two techniques are performed by constantly recalculating each cluster's center $v_i$ using Equations 64 and 65 respectively. Each data point $x_i$ is then assigned to the cluster center which is minimum against all others. Additionally, $c_i$ represents the number of data points in cluster $i$ and $w$ is a weight for improving the possible placements. Density-based is completed solely on the density of the data where so many points must fit within a specified radius, and then ambulances are assigned to each of the resulting clusters. Directly-reachable extends the idea of placement near core points by adding a threshold radius that must be considered for surrounding points. The results, as shown in Table \ref{clustering_tab_results}, showed that the Density-Based algorithm reduced the average distance up to 50\%; proving the most effective of all the selections.

\begin{equation}
    v_{i} = \frac{1}{c_i}\sum_{j=1}^{c_i} x_i
\end{equation}
    
\begin{equation}
    v_{i} = \frac{1}{\sum w_i}\sum_{j=1}^{c_i} x_i w_i
\end{equation}

\begin{table}[!ht]
\caption{Comparison of results for each clustering method.}
\label{clustering_tab_results}
\resizebox{0.6\columnwidth}{!}{\begin{tabular}{|l|l|}
\hline
\textbf{Algorithm} & \textbf{Average Distance Travelled by all Ambulances (m)} \\ \hline
Current            & 6,560                                                     \\ \hline
Weighted K-Mean    & 3,166                                                     \\ \hline
Density-Based      & 2,803                                                     \\ \hline
Directly Reachable & 4,056                                                     \\ \hline
\end{tabular}}
\end{table}

Li et al. confronted a location-based problem by utilizing real traffic information to minimize the average traveling time to an incident \cite{Li:2015:LSA:2820783.2820876}. They incorporated times based on actual GPS data and resolved the model with a local search heuristic known as "partitioning around medoids" (PAM). A medoid is an object within a cluster that has a minimal dissimilarity to others around it.  Algorithm \ref{pam_code} presents this solution where the input is a road network $G$, a set of emergency requests $R$, a travel time matrix $M$, and an initial set of $k$ ambulance stations $F_{ini}$. The output of the algorithm is a set of $k$ ambulance stations. In it each current medoid $p$ was replaced by every vertex $p'$, and the average travel time of the replacement was estimated (lines 4 to 12). Following the replacement, the one with the largest reduction is maintained forward (line 14). Li et al. further improved the Algorithm \ref{pam_code} by selecting $k$ initial stations, then pruning unpromising vertex replacements. This was performed until the travel time could not be further reduced. The results of this solution minimized travel time by 29.9\% compared to the original locations that were in use.

\medskip
\begin{algorithm}[H]
 $bsf = \sum_{r \in R}tF_{ini}(r)$\;
 $F_{ini} = F_{bsf}$\;
 \While{travel time can be reduced}{
     \For{each $p \in F_{ini}$} {
        \For{each $p' \in V$ } {
            $F=F_{ini}-p+p'$\;
            \If{$\sum_{r \in R}t_F(r)<bsf$} {
                $bsf = \sum_{r \in R} t_F(r)$\;
                $F_{bsf} = F$\;
             }
        }
     }
     \If{$F_{ini} != F_{bsf}$} {
        $F_{ini} = F_{bsf}$\;
     }
 }
 \caption{PAM-based Refinement Algorithm}
 \label{pam_code}
\end{algorithm}
\medskip

\section{Future Research}
This area is not nearly as well explored as the VRP or MCP. While there are a number of current methodologies being tested, the research appears concentrated on simulation and well-known methods. Additionally, a significant amount of the current released works have not explored beyond their initial conference publishing. Significant real-world data is available through organizations like Ornge, yet their implemented research is locked at long running ILP solutions \cite{doi:10.1287/inte.2013.0683}. In this case there have been contributions through studies like that by Pond et al. \cite{10.1007/978-3-030-04070-3_17}. This example is still preliminary and similar to other related literature has room for improvement. The following summarizes some of the future research possibilities:
\begin{itemize}
  \item Model and solve dynamic real-time variations
  \item Hybridize algorithms to achieve unique and powerful methods
  \item Restructure and solve with deep machine learning
\end{itemize}

As already suggested throughout this survey, there is significant research in static variations of these problems. While this research is quite significant, most active systems must consider dynamic changes in the actual implementation. Static problems are more acceptable in the case of ALP, yet are difficult for ARP as diversion of vehicles or unexpected incidents are a regular occurrence. This opens the possibility for further research into dynamic scheduling and possibly even locating shifting for vehicles. Some research has suggested further building off the advantage of the parallelism seen in evolutionary algorithms, as these types of techniques function better in a dynamic shifting environment \cite{SABAR20191018}. These methods can be modified to become self-adaptive, whereby a set of configurations with solution parameters are encoded into the individual solutions of the dynamic problem. Another possibility is modifying the search space from discrete to continuous \cite{OKULEWICZ201944}. The theory is that this transfer allows the methods used to better mimic manual requests and it is noted that prediction based methods have already been applied to continuous dynamic problems. In the case of the VRP, current research has shown an improvement in accuracy when compared against variants using a discrete encoding.

There are multiple references to both VRP and MCP, since they have a very similar correspondence to ARP and ALP. In general there are only small modifications done to the objective functions and the introduction of additional constraints. This allows for the transference of existing algorithms from one to the other and the likelihood of being able to complete research using these techniques. These method have already been proven in this domain and with minor alterations there is little reason not to apply them to these instances. Additionally, there is a lack of hybrid methodologies in the current research. This type of optimization uses either multiple algorithms simultaneously for solving a problem or combining multiple algorithms together to utilize the advantage of either individually \cite{YANG2014213}. Admittedly, there is a difficulty with successfully executing this as hybridizing usually requires a substantial understanding of the respective algorithms. Even with detailed knowledge, there is still a level of randomization and trial and error. Many algorithms may be incompatible with each other, while simpler methods like local search may be quite easy to join to more complex versions. This has been already applied successfully to problems like the VRP \cite{10.1007/978-3-642-28942-2_19}. However, even in this case the research is still very preliminary at best and limited to evolutionary approaches.

A somewhat surprising missing link in the list of techniques are those applied to the domain of machine learning. In general, neural network-based methods are not appropriate for all optimization problems; although, the lack of research in this area does offer a great deal of opportunity. One method seeing some gradual introduction is around reinforcement learning. Other similar domain methods, like supervised learning require a set of labelled training data to be utilized successfully. This may be difficult to achieve in certain cases as the exact solutions needed for training may be unknown and difficult to achieve. Reinforcement learning resolves this issues by eliminating the requirement of a labelled set. This area uses a state-action pair, where the goal is to train a model to take actions which maximize a reward. This can almost be seen as a sort of game as the system is trained to take the best possible move-set in order to be successful \cite{Kaelbling:1996:RLS:1622737.1622748}. This is not necessarily a new area; however, recent research has built upon this concept with the introduction of deep reinforcement learning \cite{Mnih2013PlayingAW}. Using deep learning techniques like convolutional neural networks and combining them with algorithms like Q-learning, extremely difficult problems can be solved. Two issues with this method are the timing required for training and the formulation of the states to actions. Machine learning models are notorious for taking a significant amount of time to train and with changing requirements, this may make it difficult to maintain an acceptable model. On the other hand, the input and output will need to be encoded properly to be able to properly implement this technique. A great deal of understanding will be needed for the problem space and the data must be designed in a way to ensure the results are meaningful. Research in this area is not without a starting point, and there has been some work completed related to the VRP and other combinatorial optimization problems \cite{deep, Dai:2017:LCO:3295222.3295382}.

\section{Conclusion}
The design of EMS systems are incredibly complicated, yet very important given the critical nature of their environment. Both location and routing problems have their own respective issues and vary greatly in design. Primary focus has been towards conferences, which restricts more advanced findings and limits a great deal of this research to preliminary steps. Additionally, both of these problems are NP-hard and utilize similar meta-heuristics for their solutions. As seen in Table~\ref{tab:works_arp} and Table \ref{tab:works_alp}, much of the work has been completed on static variations of the problem with heavy emphasis on simulation or biological algorithms. In the case of the former, there is a severe limitation on this as simulation does not necessarily lead to the best possible solution. Its primary purpose is to emulate an existing or theoretical system, making true optimization a problem and generalization not a guarantee. This is not to suggest simulation is useless, and to the contrary, may show faults within an existing system. Rather than being utilized solely, it should be viewed as an initial step before actual algorithms are introduced for improvement. Overall, there is still significant research required to ensure optimal EMS systems, with current literature providing ground work for advancement and comparison.


\bibliographystyle{unsrt}  
\bibliography{references}

\end{document}